\documentclass{article}
\usepackage{amssymb}

\usepackage[preprint]{corl_2025} 

\usepackage{cite}
\usepackage{graphicx}
\usepackage{amsmath,amssymb,amsfonts}
\usepackage{listings}
\usepackage{algorithmic}
\usepackage{textcomp}
\usepackage{xcolor}
\usepackage{enumitem} 
\usepackage{wrapfig} 
\usepackage{subcaption}

\usepackage{dirtytalk}
\usepackage{listings}
\usepackage{booktabs}
\usepackage{makecell}  
\usepackage{multirow}
\usepackage{placeins}

\lstdefinestyle{pythonstyle}{
    language=Python,
    basicstyle=\ttfamily\tiny,
    keywordstyle=\color{blue}\bfseries,
    commentstyle=\color{green!50!black},
    numbers=none,
    numberstyle=\tiny,
    stepnumber=1,
    breaklines=true,
    frame=single,
    backgroundcolor=\color{gray!10},
    tabsize=4,
    showspaces=false,
    showstringspaces=false
}

\lstdefinestyle{pythonhighlightstyle}{
    language=Python,
    basicstyle=\ttfamily\tiny,
    keywordstyle=\color{blue}\bfseries,
    commentstyle=\color{green!50!black},
    breaklines=true,
    numbers=none,
    frame=single,
    backgroundcolor=\color{gray!10},
    tabsize=4,
    showspaces=false,
    showstringspaces=false,
}

\lstdefinestyle{vartextstyle}{
    language={},
    basicstyle=\ttfamily\tiny,
    breaklines=true,
    numbers=none,
    frame=single,
    backgroundcolor=\color{gray!10},
    tabsize=4,
    showspaces=false,
    showstringspaces=false,
    moredelim=[s][\colorbox{yellow}]{\{}{\}}
}
\lstdefinestyle{plaintextstyle}{
    language={},
    basicstyle=\ttfamily\tiny,
    breaklines=true,
    numbers=none,
    frame=single,
    backgroundcolor=\color{gray!10},
    tabsize=4,
    showspaces=false,
    showstringspaces=false,
}

\usepackage{graphicx,booktabs,makecell}

\usepackage{etoolbox} 
\usepackage[skip=4pt]{caption} 
\usepackage[utf8]{inputenc} 
\usepackage[T1]{fontenc}
\usepackage{textcomp}

\usepackage{siunitx} 

\usepackage{listings}
\usepackage{listingsutf8} 

\title{Text2Touch: Tactile In-Hand Manipulation with LLM-Designed Reward Functions}

\author{
Harrison Field, Max Yang, Yijiong Lin, Efi Psomopoulou, David Barton, Nathan F. Lepora \\
$^1$School of Computer Science, University of Bristol, $^2$Bristol Robotics Laboratory\\  
$^3$School of Engineering Mathematics and Technology, University of Bristol\\
\texttt{\{harry.field, n.lepora\}@bristol.ac.uk} \\
}

\begin{document}

\maketitle


\begin{abstract}
Large language models (LLMs) are beginning to automate reward design for dexterous manipulation. However, no prior work has considered tactile sensing, which is known to be critical for human-like dexterity. We present \texttt{Text2Touch}, bringing LLM-crafted rewards to the challenging task of multi-axis in-hand object rotation with real-world vision based tactile sensing in palm-up and palm-down configurations. Our prompt engineering strategy scales to over 70 environment variables, and sim-to-real distillation enables successful policy transfer to a tactile-enabled fully actuated four-fingered dexterous robot hand. \texttt{Text2Touch} significantly outperforms a carefully tuned human-engineered baseline, demonstrating superior rotation speed and stability while relying on reward functions that are an order of magnitude shorter and simpler. These results illustrate how LLM-designed rewards can significantly reduce the time from concept to deployable dexterous tactile skills, supporting more rapid and scalable multimodal robot learning.\\ Project website:\texttt{ \href{https://hpfield.github.io/text2touch-website/}{https://hpfield.github.io/text2touch-website/}}
\end{abstract}



\begin{wrapfigure}{R}{0.6\textwidth}
    \vspace{-2em}  
    \centering
    \includegraphics[width=0.58\textwidth]{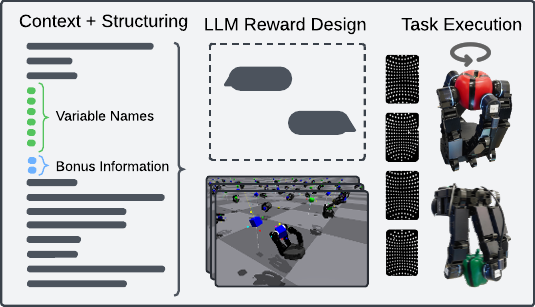}
    \caption{ \texttt{Text2Touch} improves upon previous reward function design methods to increase the performance of robotic in-hand object rotation in rotation speed and grasp stability. We evaluate the performance of LLM-generated reward functions using only tactile and proprioceptive information in the real world.}
    \label{fig:splash}
    \vspace{-10pt} 
\end{wrapfigure}

\section{Introduction}
Designing reinforcement learning (RL) reward functions for dexterous in-hand manipulation remains a formidable challenge. Traditional approaches often rely on domain experts to painstakingly specify and tune reward terms \citep{kober2013reinforcement}, a process prone to suboptimal or unintended behaviours \citep{hadfield2017inverse, booth2023perils}. Recent work has shown that large language models (LLMs) can generate policy or reward code for robotic tasks \citep{liang2023code, eureka, yu2023language, xie2023text2reward, guo2024utilizing, turcato2025towards}, a notable step toward reducing manual engineering. However, these breakthroughs have primarily focused on conventional sensing modalities (vision, proprioception) for real-world validation \citep{liang2023code, dreureka, yu2023language, xie2023text2reward, guo2024utilizing}. To date, tactile sensing has not yet been integrated into automated reward generation via LLMs in either simulated or real-world settings.

Vision based tactile sensing can provide detailed contact and force signals that visual sensing alone often fails to capture, especially under occlusions or subtle slip conditions \citep{hansen2022visuotactile}. By detecting fine-grained grasp instabilities and contact dynamics, vision based tactile sensors have enabled robots to perform dexterous actions such as in-hand object rotation \citep{anyrotate, qi2023general, qi2023hand}. Yet incorporating complex sensor modalities like touch into learned control systems usually amplifies the difficulties of reward design, especially in the context of high-DoF systems like fully-actuated multi-fingered hands \citep{kober2013reinforcement, russell2016artificial}. Given this, automatically generating effective tactile-based rewards may be key to unlocking robust policies for advanced manipulation. In this work we tackle gravity-invariant, multi-axis in-hand object rotation: the robot must keep the object suspended off the palm while rotating it around 3 perpendicular axes under both palm-up and palm-down configurations, providing sufficient palm orientations to demonstrate the adaptive capabilities of trained policies. 

In this paper, we investigate whether LLMs can facilitate reward function design for real-world in-hand manipulation using tactile feedback. We build on previous iterative LLM reward design work \citep{eureka}, which demonstrated that GPT-4 could automatically produce reward code that outperformed human-crafted counterparts in multiple simulated tasks. However, those experiments (like most related efforts \citep{liang2023code, dreureka}) did not include tactile inputs. Our approach, \texttt{Text2Touch}, extends LLM-based reward generation to cover tactile sensor data, then applies a teacher–student pipeline for sim-to-real transfer. We show that, with careful prompt engineering and environment-context provision, LLMs can autonomously design concise, interpretable reward functions that drive successful in-hand rotation policies on a real tactile Allegro Hand, a fully actuated four-fingered dexterous robot hand outfitted with vision-based TacTip tactile sensors. By reducing reliance on expert tuning, \texttt{Text2Touch} highlights the promise of LLMs as a practical tool for accelerating multi-modal RL in complex physical settings.

\noindent \textbf{Our main contributions are:} \vspace{-0.4em}
\begin{enumerate}[leftmargin=*,itemsep=0pt,topsep=0pt]
    \item \textbf{Novel LLM-based reward design for tactile manipulation.} 
    We introduce the first automated reward-generation approach to incorporate tactile sensing, bridging a gap in existing LLM-based robotics. \vspace{-0.35em}
    \item \textbf{Enhanced prompting for high-complexity environments.} 
    We propose a refined prompt-structuring strategy that scales to over 70 environment variables, greatly reducing code failures and improving simulation performance. \vspace{-0.35em}
    \item \textbf{Real-world validation via sim-to-real distillation.} 
    We transfer an LLM-designed policy to a tactile Allegro Hand that rotates an aloft object about all three axes in both palm-up and palm-down configurations, surpassing the best published results (human-engineered) on this state-of-the-art dexterity task. \vspace{-0.35em}
\end{enumerate}
\vspace{-0.1em}

\section{Related Work}
\vspace{-0.3em}
\paragraph{LLMs for Reward and Policy Design.}
 Several works have recently employed LLMs to alleviate the burden of reward engineering. Eureka \citep{eureka} introduced an iterative pipeline where GPT-4 generates and refines reward functions through evolutionary search, outperforming handcrafted baselines in simulation. Further developments incorporated pre-trained language models to translate user-provided descriptions into Python reward snippets, enabling quadruped and manipulator control \citep{Language-to-rewards-for-robotic-skill-synthesis}. A similar direction was explored by \citet{sarukkai2024automated} and \citet{xie2024large} in developing LLM-driven progress estimators and white-box reward decomposition. Despite promising results, most studies operated in controlled simulators or used only proprioception and vision in the real world \citep{dreureka, yu2023language, xie2023text2reward, guo2024utilizing}, none addressed tasks requiring high-dimensional, vision based tactile sensing.

\paragraph{Related LLM-Based Methods for Robotic Control.}
Beyond reward and policy design, LLMs have been used to generate high-level robot plans \citep{saycan2022arxiv, liang2023code, mon2025embodied}, directly produce policy code \citep{liang2023code}, and unify multimodal sensor inputs \citep{driess2023palm}. However, these approaches often assume access to pre-trained skills \citep{saycan2022arxiv, mon2025embodied} or rely on computationally expensive vision encoders \citep{driess2023palm, brohan2022rt, brohan2023rt}, whereas we focus on training novel tactile-based skills without large-scale pretrained policies. Similarly, although automatic curriculum design and environment generation methods \citep{ryu2024curricullm, liang2024environment} can help robots acquire complex skills, we intentionally employ a fixed, human-designed curriculum in this project to isolate the effects of our LLM-generated rewards and directly compare them against a human-engineered baseline. Our approach centres on specifying robust reward functions adapted to tactile sensing directly via LLM-enabled teacher agents, filling a gap in the literature where prior methods rarely address tactile feedback’s unique challenges.

\paragraph{Advances in Tactile Robotic Manipulation.}
Modern sensor technologies such as GelSight, DIGIT and TacTip capture high-resolution images of contact patches \citep{hansen2022visuotactile, yuan2017gelsight, lambeta2020digit, lepora2021soft, lin2022tactilegym2}, affording information on slip, shear forces and object shape. Recent work shows that such tactile cues boost grasp stability and in-hand dexterity compared with vision or proprioception alone \citep{hansen2022visuotactile, lin2023bi, james2020slip, borntoengineer2023rotatinghand}. For example, \citep{anyrotate} not only demonstrated the first gravity-invariant, multi-axis object rotations in simulation and hardware, but also showed that achieving this level of dexterity critically depended on rich tactile sensing, a capability that earlier works, which managed only simpler tasks like single-axis rotation or static palm balancing, lacked. Yet these successes still rely on carefully hand-tuned reward functions, a process that is slow, demands expert intuition (which is biased by human assumptions) and must be repeated whenever the task, object set or sensor suite changes.

\paragraph{Sim-to-Real Transfer with LLM-Generated Rewards.}
Although sim-to-real adaptation via domain randomisation or teacher–student strategies has seen progress (\citep{andrychowicz2020learning, peng2018sim, akkaya2019solving, chebotar2019closing, chen2023visual, chen2024general}), it remains unclear whether automatically generated reward structures carry over to hardware for tactile tasks. Early steps toward real-world LLM-based reward design were demonstrated in \citep{dreureka}, but limited to proprioceptive sensing. In contrast, we extend automatic reward design to a tactile modality, scale it to an environment exposing over seventy state variables, and demonstrate transfer on a tactile Allegro Hand using a bespoke LLM reward design framework and a teacher–student pipeline.

\begin{figure}[t]
    \centering
    \includegraphics[width=\textwidth]{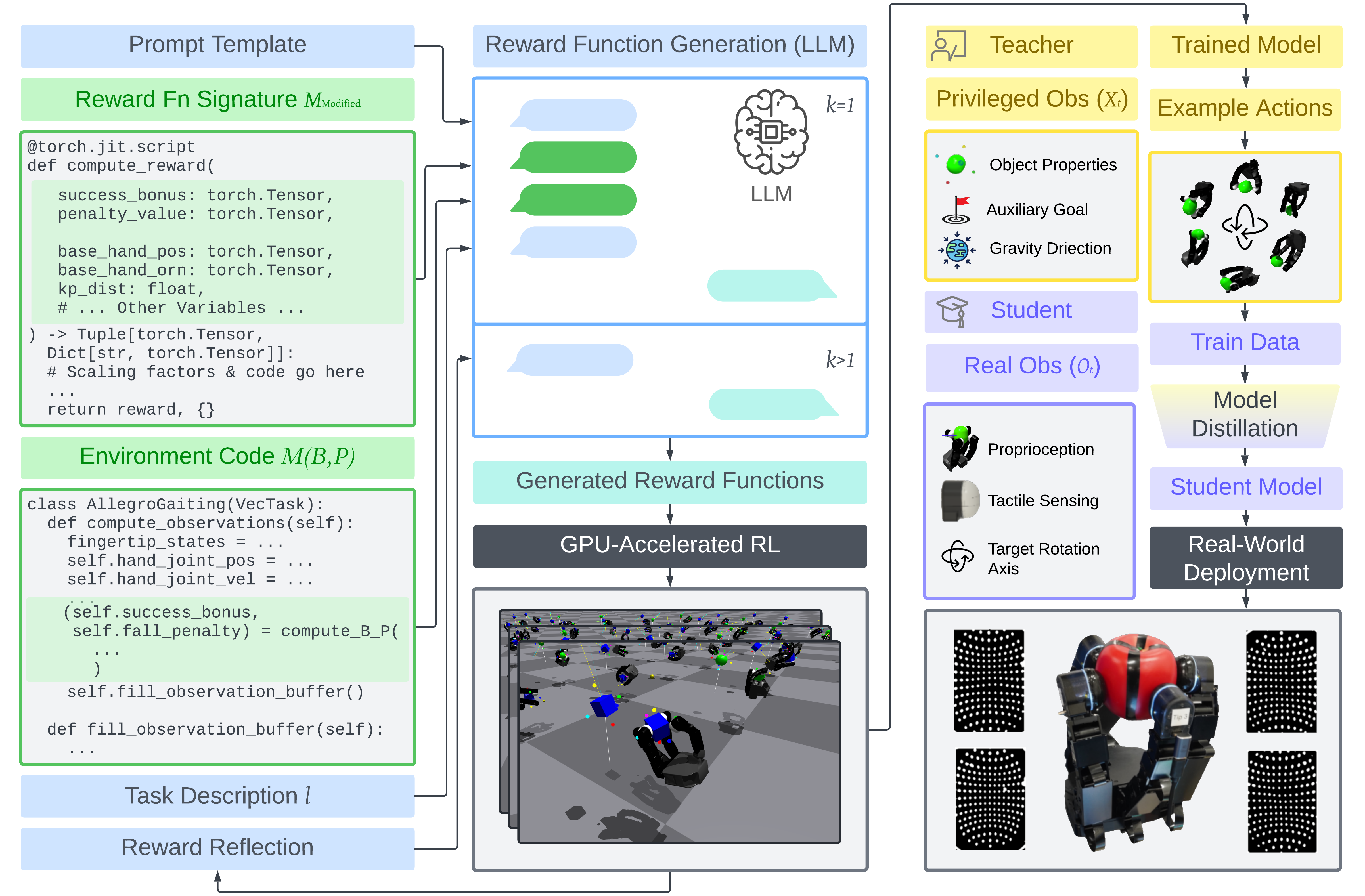}
    \caption{ \texttt{Text2Touch} training and deployment pipeline. The left and middle columns comprise the reward generation pipeline with our novel prompting strategy components in green, and the right column denoting the model distillation and deployment phase. The teacher model (yellow) is the final output of the reward generation phase and the student (lilac) is the distilled teacher using only real-world observations.}
    \label{fig:pipeline}
    \vspace{-1.5em}
\end{figure}

\section{Methods} \label{sec:methods}

We introduce \texttt{Text2Touch}, a framework enabling LLMs to generate reward functions through advanced prompt strategies, endowing the model with new abilities (e.g. scalable bonuses and penalties) and a refined information structure so that desired behaviours can evolve over successive iterations. Beginning with a natural language task description and an environment context, these prompts guide the LLM to propose, evaluate and iteratively refine reward candidates that elicit precise rotational behaviour in simulation. The optimal reward is then distilled into a policy conditioned solely on proprioceptive and tactile inputs, validated in the real world on a tactile Allegro Hand.

\subsection{Iterative LLM Reward Design}
\label{eureka-reward-design}

We build upon the Eureka framework~\citep{eureka} to automate reward design, using a PPO learning agent~\citep{schulman2017proximal} for policy optimisation. Given environment code/context \(M\) and a natural-language task description \(l\), we prompt an LLM to generate reward functions \(R\) that maximise task score \(s\). At each iteration \(k\), a batch of candidates \(\{R^{(k)}\}\) is evaluated by training policies \(\{\pi^{(k)}\}\) under a fitness function \(F\). The best reward \(R_{\text{best}}^{(k)}\) is refined through \emph{reward reflection} (Appendix \ref{app:prompting-strategies}) to produce \(\{R^{(k+1)}\}\). This loop, represented in \eqref{eq:pipeline} and Figure \ref{fig:pipeline} Left \& Middle, repeats until a preset limit.

\paragraph{Natural Language Task Description.} We define the natural language task description $l$ as: ``To imbue the agent with the ability to reposition and reorient objects to a target position and orientation by regrasping or finger gaiting, where contacts with the object must be detached and repositioned locally during manipulation." This description was selected after experiments showed simpler prompts lacked sufficient clarity, while overly detailed prompts introduced undesirable human biases. Our chosen formulation closely aligns with the authors description of the human-engineered baseline task, precisely delineating task objectives without imposing qualitative biases.

\paragraph{Environment-Based Subgoal Curriculum.}
The task is to rotate an object about a target axis. Each time a \textit{sub-goal incremental rotation} is met, the environment updates the target orientation and continues the episode, allowing multiple successive ``successes'' per rollout (Appendix \ref{app:model-training}). We track these as a running \texttt{successes} count, which forms the fitness \(F\). A final termination condition resets the episode if time expires (600 simulation steps equating to 30s) or the agent deviates excessively.

\begin{wrapfigure}{r}{0.6\textwidth}
  \vspace{-1em} 
  \begin{minipage}{\linewidth}
    \begin{equation}\label{eq:pipeline}
      \begin{aligned}
        &\text{Initialization:}\quad \{R^{(0)}\} \sim \mathrm{LLM}(l, M), \\
        &\text{For each iteration } k \ge 0 : \\
        &\quad \{\pi^{(k)}\} \leftarrow \{\mathrm{RL}(\{R^{(k)}\})\}, \\
        &\quad \{s^{(k)}\} \leftarrow \{F(\{\pi^{(k)}\})\}, \\
        &\quad R_{\text{best}}^{(k)} = \mathrm{SelectBest}(\{R^{(k)}, s^{(k)}\}), \\
        &\quad \{R^{(k+1)}\} \sim \mathrm{LLM}\bigl(l, M,\mathrm{Feedback}(s_{\text{best}}^{(k)}, R_{\text{best}}^{(k)})\bigr).
      \end{aligned}
    \end{equation}
  \end{minipage}
\end{wrapfigure}

\noindent \textbf{Environment Complexity}. As the number of variables grows, the LLM must reason over a context window where individual tokens have multiple, interdependent meanings. Increased token and variable counts lead to more complex attention relationships, making it harder for LLMs to generate meaningful responses, measured here by the agent’s task success score $s$ when trained with the LLM's reward function. To address this, we propose several prompting strategies tailored for complex, variable-dense environments.

\subsection{Scaling Bonuses and Penalties}\label{scaling-bonuses}

Originally, Eureka \citep{eureka} added a fixed scalar success bonus \(B\) to the LLM-generated reward \(R_{\text{LLM}}(\mathbf{o})\),
\[
\mathcal{R}_{\text{total}} = R_{\text{LLM}}(\mathbf{o}) + B.
\]
Because the scale of \(R_{\text{LLM}}\) varies widely, a single fixed \(B\) can hamper policy learning. To address this, we instead include both the success bonus \(B\) and an early-termination penalty \(P\) in the LLM's context as scalable variables (Figure \ref{fig:pipeline} Left) without revealing the success/failure conditions,
\[
R_{\text{LLM}}^{(B,P)} \;\sim\; \text{LLM}\bigl(l, M(B,P)\bigr),
\quad
\mathcal{R}_{\text{total}} \;=\; R_{\text{LLM}}^{(B,P)}(\mathbf{o}).
\]
This lets the LLM appropriately scale \(B\) and \(P\) within its final reward expression, yielding higher task scores. Empirically, Table~\ref{tab:llm_performance} shows task success through LLM-generated rewards proved impossible without providing a scalable \(B, P\), with reward function examples shown in Appendix \ref{app:reward-functions}.

\subsection{Modified Prompt Structuring}\label{modified-prompting}

When applying the original prompt structure from Eureka \citep{eureka} to our problem and environment, we found LLMs frequently produced syntactic errors or mismatched variable types. Our environment contains over 70 variables, far more than the original Eureka examples, which are limited to $\sim$10. To mitigate this, we now supply each LLM with an explicit reward function signature containing all variable names/types (see Figure \ref{fig:pipeline} Left or Appendix \ref{app:prompting-strategies} for examples), instead of the generic signature used originally, so
\[
M_{\text{modified}} = \{ E,\; S_{\text{detailed}} \}, 
\]
where \(E\) is the environment code and \(S_{\text{detailed}}\) is an example reward function signature containing all available observations. This explicit listing reduces confusion, enabling the LLM to correctly reference variables and produce valid reward functions. In practice, it greatly decreased code failures and improved policy performance.

\subsection{Integration with Sim-To-Real Pipeline}\label{eureka-anyrotate-integration}

After generating candidate rewards via iterative LLM reward design, we integrate the trained policies into a sim-to-real pipeline. We train a Teacher policy in simulation using privileged information, then distil that policy to a Student with only tactile and proprioceptive observations (Section~\ref{sec:sim-to-real}).

\begin{figure}[t]
    \centering
    \includegraphics[width=\textwidth]{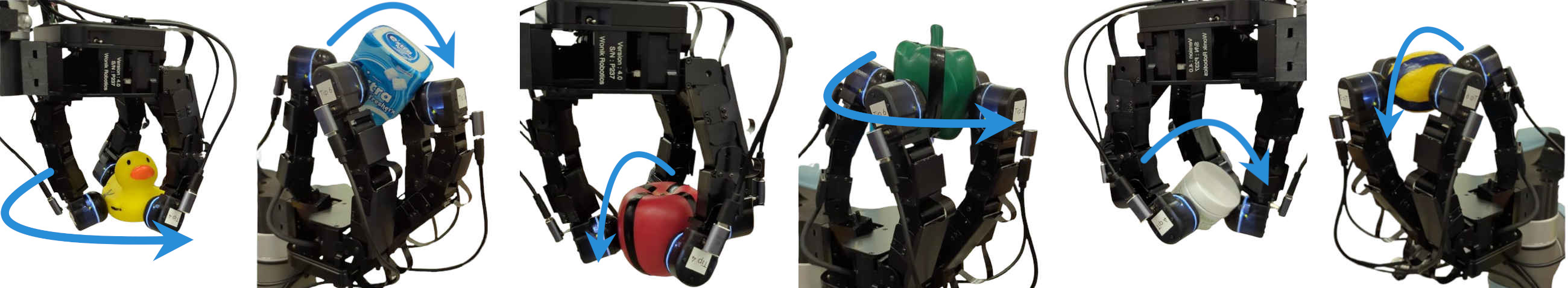}
    
    \caption{ Real-world deployment using various objects in palm-up \& palm-down configurations. Video in supplementary materials and available on the \href{https://hpfield.github.io/text2touch-website/}{project website}.}
    \label{fig:real-demos}
    \vspace{-1.5em}
\end{figure}

\paragraph{Reward Discovery Training.}
Evaluating many reward candidates with the full 8~billion-step training is  infeasible with available compute resources (Appendix \ref{app:compute-resources}). We thus use a shorter, 150~million-step run each iteration to see if a reward $R^{(k)}$ can guide policies to partial success. Concretely:
\begin{enumerate}[leftmargin=*,itemsep=0pt,topsep=3pt]
\vspace{-0.5em}
\item \textbf{Generation:} The LLM produces $\{R^{(k)}\}$ from $l, M(B,P)_{\text{modified}}$ ($k=1, \dots , 5$).
\item \textbf{Training:} Simultaneously train policies $\pi_i^{(k)}$ ($i = 1,\dots,4$) using 150~million simulation steps.
\item \textbf{Evaluation:} Compute $s_i^{(k)} = F(\pi_i^{(k)})$, based on sub-goal successes.
\item \textbf{Selection:} Choose $R_{\text{best}}^{(k)}$ and provide feedback to the LLM for the next iteration.
\end{enumerate}
Repeated short runs provide timely feedback, enabling the LLM to refine rewards and converge to a high-quality $R^*$. Further details of reward discovery training parameters in Appendix \ref{app:teacher-student}. 

\subsection{Policy transfer from simulation to hardware}\label{sec:sim-to-real}

After reward discovery we train a Teacher policy in simulation with privileged observations such as object pose and velocity.  We record roll-outs and fit a Student that observes only proprioception and tactile information, minimising the mean-squared distance to the teacher’s action at each step. Similar teacher-student paradigms have been used successfully in dexterous manipulation and sim to real transfer \citep{lee2020learning, anyrotate, qi2023general, qi2023hand, chen2020learning}.

The trained student transfers to hardware without further tuning.  An Allegro Hand equipped with TacTip vision based tactile sensors runs at 20Hz, receiving joint positions, velocities and the tactile observations. Trials cover ten household objects, three rotation axes and palm up/palm down hand orientations.  Performance is measured by complete rotations per episode and time to termination. Object dimensions and masses appear in Appendix \ref{app:real-world-objects}.

\section{Experiments and Results} \label{sec:result} \vspace{-0.5em}

We evaluate our approach in three main stages. \textbf{Stage~1} justifies the proposed prompting strategy and explores how LLMs generate reward functions under privileged information in simulation, ultimately selecting a ``best'' generated reward function for full model training per LLM. \textbf{Stage~2} distills that model with a teacher--student pipeline into a model that uses only proprioceptive and tactile inputs. In \textbf{Stage 3}, we deploy the  distilled policy on a tactile Allegro Hand. For fair comparison to the Baseline, we freeze all components of stages 2 and 3 as presented in AnyRotate \citep{anyrotate}.

\begin{table}[t]
    \centering
    \scriptsize
    \begingroup
    \renewcommand{\arraystretch}{1.15} 
    \resizebox{1.0\linewidth}{!}{
    \begin{tabular}{@{}l *{5}{ccc}@{}}
        \toprule
        \multirow{3}{*}{\makecell[c]{\\\textbf{Prompting} \\ \textbf{Strategy}}}
        & \multicolumn{3}{c}{\textbf{GPT-4o}} & \multicolumn{3}{c}{\textbf{o3-mini}}
        & \multicolumn{3}{c}{\textbf{Gemini-1.5-Flash}} & \multicolumn{3}{c}{\textbf{Llama3.1-405B}}
        & \multicolumn{3}{c}{\textbf{Deepseek-R1-671B}} \\
        \cmidrule(lr){2-4} \cmidrule(lr){5-7} \cmidrule(lr){8-10} \cmidrule(lr){11-13} \cmidrule(lr){14-16}
        & \multicolumn{2}{c}{Rots/Ep} & Solve
        & \multicolumn{2}{c}{Rots/Ep} & Solve
        & \multicolumn{2}{c}{Rots/Ep} & Solve
        & \multicolumn{2}{c}{Rots/Ep} & Solve
        & \multicolumn{2}{c}{Rots/Ep} & Solve \\
        & Best & Avg & Rate & Best & Avg & Rate & Best & Avg & Rate & Best & Avg & Rate & Best & Avg & Rate \\
        \midrule
        \makecell[l]{Bonus/Penalty+ Mod}
            & \textbf{5.46} & 5.34  & \textbf{84\%}
            & \textbf{5.38} & 5.26 & \textbf{28\%}
            & \textbf{5.48} & 5.29 & \textbf{31\%}
            & 5.41 & 5.28 & \textbf{10\%}
            & 5.08 & 5.03 & \textbf{16\%} \\
        \makecell[l]{Bonus/Penalty}
            & 0.10 & 0.09 & 0\%
            & 0.17 & 0.17 & 0\%
            & 0.04 & 0.02 & 0\%
            & \textbf{5.42} & 5.23 & \textbf{10\%}
            & \textbf{5.26} & 5.12 & \textbf{16\%} \\
        \makecell[l]{Mod template}
            & 0.17 & 0.16 & 0\%
            & 0.17 & 0.12 & 0\%
            & 0.10 & 0.09 & 0\%
            & 0.02 & 0.01 & 0\%
            & 0.16 & 0.16 & 0\% \\
        \makecell[l]{Original}
            & 0.17 & 0.15 & 0\%
            & 0.18 & 0.14 & 0\%
            & 0.17 & 0.17 & 0\%
            & 0.18 & 0.13 & 0\%
            & 0.15 & 0.13 & 0\% \\
        \bottomrule
        \vspace{0.1em}
    \end{tabular}%
    
    }
    
    \endgroup

    \caption{Highest $360^{\circ}$ rotations per episode (Rots/Ep) split into best, average and overall frequency of episodes with $s>1$ (Solve Rate) across 5 LLMs and 4 prompting strategies.}
    \label{tab:llm_performance}
    \vspace{-2.5em}
\end{table}

\subsection{Stage 1: Prompting Strategy Verification and Privileged Information Model Training}

\begin{table}[b]
\vspace{-0.5em}
\centering
\scriptsize
\begin{tabular}{lrrrrrrrrrrrrrr}
\toprule
& \multicolumn{6}{c}{\textbf{Performance}} & \multicolumn{6}{c}{\textbf{Code Quality}} \\
\cmidrule(lr){2-7}  \cmidrule(lr){8-13} 
\multirow{2}{*}{\textbf{LLM}} & \multicolumn{2}{c}{\textbf{Rots/Ep ↑}} & \multicolumn{2}{c}{\textbf{EpLen(s) ↑}} & \textbf{Corr} & \textbf{GR ↑} & \multicolumn{2}{c}{\textbf{Vars ↓}} & \multicolumn{2}{c}{\textbf{LoC ↓}} & \multicolumn{2}{c}{\textbf{HV ↓}}   \\
\cmidrule(lr){2-3} \cmidrule(lr){4-5} \cmidrule(lr){6-6} \cmidrule(lr){7-7} \cmidrule(lr){8-9} \cmidrule(lr){10-11} \cmidrule(lr){12-13}
 & Best ($\pi^{*}$) & Avg & $\pi^{*}$ & Avg & $\pi^{*}$ & Avg & $\pi^{*}$ & Avg & $\pi^{*}$ & Avg & $\pi^{*}$ & Avg  \\
\midrule
        Baseline & 4.92 & 4.73 & \textbf{27.2} & \textbf{26.8} & 1 & - & 66 & - & 111 & - & 2576 & - \\
        \midrule
        Gemini‑1.5-Flash & \textbf{5.48} & \textbf{5.29} & 24.1 & 23.8 & 0.40 & 0.76 & 7 & 6.3 & 24 & 22.6 & 370 & 301  \\
        Llama3.1-405B & 5.41 & 5.28 & 23.7 & 23.2 & 0.35 & 0.45 & 5 & 6.6 & 31 & 22.5 & 211 & 233  \\
        GPT-4o & 5.46 & 5.20 & \underline{24.4} & 23.4 & 0.30 & 0.63 & 8 & 8.1 & 35 & 26.9 & 317 & 300 \\
        o3‑mini & 5.38 & 5.26 & 23.9 & 23.1 & 0.47 & 0.92 & 6 & 6.6 & 27 & 30.1 & 281 & 302  \\
        DeepseekR1-671B & 5.26 & 5.12 & 22.9 & 22.4 & 0.42 & 0.93 & 12 & 11.9 & 43 & 45.3 & 994 & 699  \\
\bottomrule
\hfill
\end{tabular}
\caption{ Comparison of LLM-generated and human-designed reward functions when used to train models with privileged information in simulation. \textbf{Performance:} Rotations per episode (Rots/Ep), episode lengths (EpLen(s)), reward correlation with baseline (Corr) and generation rate of runnable code (GR) of the model with the highest Rots/Ep ($\pi^{*}$) and the average over that reward function's repeated training runs. \textbf{Code Quality:} For the reward function producing $\pi^{*}$ and the average over all trained reward functions, we record the variables used from the environment (Vars), the lines of code in the function body (LoC) and the Halstead Volume (HV).}
\label{tab:llm-extensive-s1}
\end{table}

\noindent \textbf{Iterative Reward Generation.}
We conduct five reward discovery experiments for each prompting strategy-LLM pair listed in Table~\ref{tab:llm_performance}. Each experiment follows the iterative reward-generation loop (Section~\ref{eureka-reward-design}) for five iterations, yielding 20 reward functions per experiment. Since each prompting strategy-LLM pair has five experiments, we consider 100 generated reward functions for each cell in Table~\ref{tab:llm_performance}. Over four prompting strategies and five LLMs, this amounts to 2{,}000 total reward functions.

From these short-run experiments, we select the single best reward for each cell in Table~\ref{tab:llm_performance} and train it from scratch for 8\,billion steps, repeating this final training run five times. We then evaluate each resulting policy for 50{,}000 simulation steps and report the highest rotation score as the final entry in each cell. This procedure confirms the value and generalisability of our prompting approach \(M(B,P)_{\text{modified}}\) by comparing it against multiple prompt variations across five distinct LLMs. Of particular note is the necessity of the success bonus $B$ and fall penalty $P$ as scalable parameters to the LLM for any success, marking a significant departure from the capabilities of Eureka $M$.

\noindent \textbf{LLM Selection.} We aim for a spread of open (\citep{grattafiori2024llama}, \citep{guo2025deepseek}) vs closed (\citep{hurst2024gpt}, \citep{OpenAIo3mini}, \citep{team2024gemini}) weights, reasoning (\citep{guo2025deepseek}, \citep{OpenAIo3mini}) vs non-reasoning (\citep{grattafiori2024llama}, \citep{hurst2024gpt}, \citep{team2024gemini}) instruct-tuned models to represent the capabilities of models available at the time of conducting these experiments. We explain our selection rationale further in Appendix \ref{app:llm-commentary}, and provide supplementary experiment details testing $M(B,P)_{modified}$ on other LLMs in the Appendix \ref{app:llms}.

\noindent \textbf{Performance Across LLMs.} In Table \ref{tab:llm-extensive-s1} (Left), we find that properly prompted LLMs are consistently capable of achieving a higher number of rotations per episode than the human baseline.  The key difference between the human-engineered design process and the automated reward generation loop is that the automated loop has less contextual information than its human counterpart. The LLM must reason over a tightly constrained context of tactile and other environment variables to produce a reward function that successfully transfers to the real world. We see this constraint in the low reward correlation to the baseline and the comparably low episode lengths seen when using generated reward functions.

\noindent \textbf{Code Quality.} In Table~\ref{tab:llm-extensive-s1} (Right), we compare the code quality of LLMs with the human baseline to illustrate that our method brings improved interpretability alongside better performance. Not only do the LLM-generated rewards consistently outperform the human baseline, they also achieve this using approximately one-tenth the variables from the environment (Vars), one-quarter the lines of code in the function body (LoC), and one-eighth the Halstead Volume \citep{halstead1977elements} (HV). These reductions translate to significantly simpler reward functions that directly impact interpretability and computational cost.

\textbf{Comparison of Reward Composition Strategies.}
Across all five LLMs, we observe a common pattern of reward design built around a few succinct components for hand–object contact, object position/orientation, and sparse success bonuses. Each component is typically assigned a separate scale before combining them into a final weighted sum. This concise decomposition contrasts with the human-engineered baseline, which encodes similar ideas in more fragmented and interconnected terms (for example, multiple overlapping orientation and smoothness terms).

Among the five models, two exhibit more distinctive behaviours. o3-mini includes extensive in-line commentary connecting each component's weight choice to the observed performance, effectively explaining its design decisions after each iteration. Deepseek-R1-671B is the only model that leverages the object's keypoint data as a unified position–orientation signal, mirroring aspects of the environment's fitness function that it never directly sees, while still integrating contact bonuses. Meanwhile, Gemini-1.5-Flash produced the highest overall rotation scores by separating out multiple scales and temperatures from the final summation, letting it adapt each term’s importance non-linearly. These examples demonstrate that LLMs can both discover novel reward design ideas and distil them into concise, interpretable code. We provide examples of the highest performing reward function generated by each LLM, along with the human-engineered baseline, in Appendix \ref{app:reward-functions}. \vspace{-0.5em}

\begin{wraptable}{R}{0.6\textwidth}
    \vspace{-1em}
    \centering
    \scriptsize
    \begin{tabular}{l c c c c}
        \toprule
        \multirow{2}{*}{\textbf{LLM}} & \multicolumn{2}{c}{\textbf{{OOD Mass}}} & \multicolumn{2}{c}{\textbf{{OOD Shape}}}  \\
        & Rots/Ep ↑ & EpLen(s) ↑ & Rots/Ep ↑ & EpLen(s) ↑ \\
        \midrule
        Baseline & 2.94 & \textbf{23.0} & 2.44 & \textbf{25.1} \\
        \midrule
        Gemini-1.5-Flash & \textbf{3.38} & 19.8 & \textbf{2.68} & 21.3 \\
        GPT-4o & 3.35 & 20.7 & 2.62 & 22.5 \\
        o3-mini & 3.25 & 19.2  & 2.52 & 21.3 \\
        Llama3.1-405B & 3.02 & 18.1 & 2.50 & 20.0 \\
        Deepseek-R1-671B & 3.32  &  22.7 & 2.47  & 23.4  \\
        \bottomrule
    \end{tabular}
    \caption{ Tactile and proprioceptive observation models distilled from privileged agents ($\pi^{*}$ in Table~\ref{tab:llm-extensive-s1}). We report on average rotation achieved per episode (Rots/Ep) and average episode length (EpLen, max 30) for arbitrary rotation axis and hand direction.}
    \label{tab:llm-s2}
    \vspace{-3.1em}
\end{wraptable}

\subsection{Stage 2: Model Distillation to Tactile Observations} \vspace{-0.5em}
We distil each LLM's $\pi^{*}$ from Stage 1 into a tactile \& proprioceptive Student model as outlined in Section \ref{sec:sim-to-real}.\vspace{-1em}
\paragraph{Out-of-Distribution Testing.}

We evaluate each Student on heavier objects and novel shapes (Figure~\ref{fig:objects}, top), which were not seen during training. As shown in Table~\ref{tab:llm-s2}, Students distilled from LLM-generated rewards consistently outperform the human baseline in terms of average rotations per episode, mirroring results from Stage~1. Notably, the Deepseek-based Student exhibits longer episode durations when using tactile inputs than it did with privileged observations. This suggests that the restricted sensory modality can still produce highly effective rotation behaviours, possibly because focusing on fewer but more task-relevant signals (tactile contact) helps the policy learn robust solutions without being overwhelmed by the full state space.

\subsection{Stage 3: Real-World Comparison of Distilled Models}\label{sec:real-world} \vspace{-0.5em}

\begin{wrapfigure}{r}{0.45\textwidth}
    \vspace{-10pt}  
    \centering
    \includegraphics[width=0.43\textwidth]{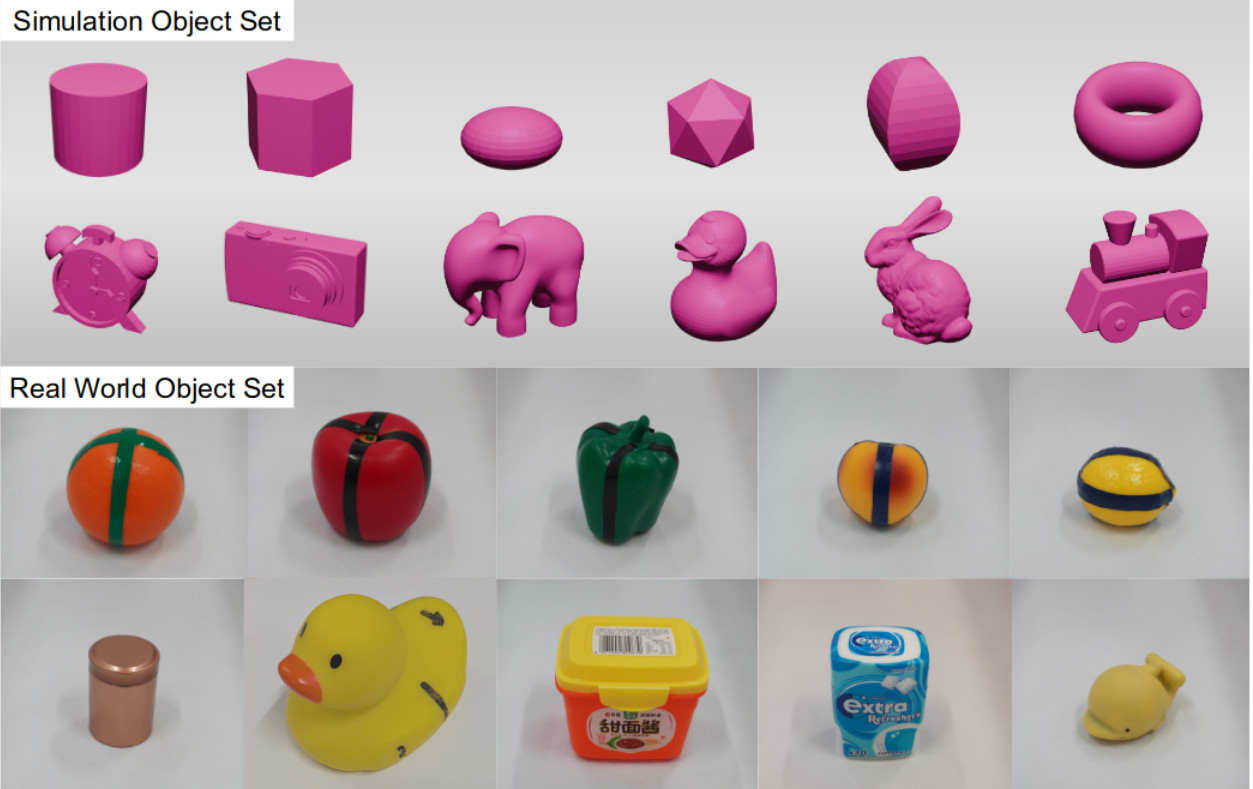}
    \caption{ Top: Objects used to evaluate distilled Student tactile model in simulation. Bottom: Objects used to evaluate distilled models in the real world.}
    \label{fig:objects}
    \vspace{-1em}
\end{wrapfigure}

We validate our distilled Student policies using the hardware setup and protocol described in Section \ref{sec:sim-to-real} with objects seen in Figure \ref{fig:objects}, bottom. \vspace{-0.5em}

\paragraph{Model Selection.}
From Stage~2, we select three LLM-based policies to test in real-world conditions alongside the human-engineered baseline. Gemini-1.5-Flash (highest rotations per episode), GPT-4o (highest solve rate), and Deepseek-R1-671B (longest episodes). Following the process outlined in stage 2, each policy is deployed in the real-world unchanged.\vspace{-0.5em}

\paragraph{Performance and Analysis.}
Table~\ref{tab:avg-orientation-axis} shows that all three LLM-based policies outperform the human-engineered baseline in average rotations per episode and also sustain longer episodes before termination. Notably, Deepseek-R1-671B achieves a \textbf{38\%} increase in rotations and a \textbf{25\%} increase in episode duration compared with the baseline. Although the baseline policy appeared more stable than the LLM-based policies in simulation, here we see the opposite: the LLM-based approaches demonstrate greater real-world stability overall.

The improved real-world stability of LLM-generated policies likely stems from differences between simulated and physical tactile sensing. Simulated contact uses rigid-body models, producing brittle signals, whereas real-world vision-based tactile sensors provide smooth, continuous feedback. LLM-trained agents, optimised for faster motions in simulation, became highly sensitive to small contact changes. Although this sensitivity sometimes caused failures in simulation, richer physical feedback enabled early instability detection and regrasping in the real world. In contrast, the human-engineered baseline, favouring slower, cautious motions, lacked this reactive advantage. This behaviour is most prominently represented rotating the Duck about the $z$ axis in a palm up orientation, which can be seen in the supplementary materials video. Based on empirical (Tables \ref{tab:llm-s2} \& \ref{tab:avg-orientation-axis}) and qualitative observations, we hypothesise that the LLM-derived policies' increased speed, initially a stability liability in simulation, was crucial to their improved real-world performance. \vspace{-0.5em}

\begin{table}[t]
\centering
\begingroup
    \setlength{\tabcolsep}{3pt}
    \scriptsize
\begin{tabular}{lcccccccccccc|cc}
\toprule
\textbf{Approach} & \multicolumn{2}{c}{\textbf{Palm Up Z}} & \multicolumn{2}{c}{\textbf{Palm Down Z}} & \multicolumn{2}{c}{\textbf{Palm Up Y}} & \multicolumn{2}{c}{\textbf{Palm Down Y}} & \multicolumn{2}{c}{\textbf{Palm Up X}} & \multicolumn{2}{c}{\textbf{Palm Down X}} & \multicolumn{2}{c}{\textbf{Total Avg}} \\
 & \textbf{Rot} & \textbf{TTT} & \textbf{Rot} & \textbf{TTT} & \textbf{Rot} & \textbf{TTT} & \textbf{Rot} & \textbf{TTT} & \textbf{Rot} & \textbf{TTT} & \textbf{Rot} & \textbf{TTT} & \textbf{Rot} & \textbf{TTT} \\
\midrule
Human-engineered Baseline & 1.42 & 22.1 & 0.96 & 17.2 & \textbf{1.04} & 23.4 & 0.67 & 24.0 & 1.23 & 19.0 & 0.65 & 14.1 & 0.99 & 20.0 \\
GPT-4o & 2.34 & 26.5 & \textbf{1.67} & \textbf{23.1} & 1.00 & \textbf{27.1} & 0.46 & 16.9 & 0.92 & 14.9 & 0.73 & 15.6 & 1.18 & 20.7 \\
Gemini-1.5-Flash & 2.12 & 26.9 & 1.19 & 19.0 & 1.00 & 26.2 & 0.61 & 27.5 & \textbf{1.47} & 21.4 & \textbf{1.31} & \textbf{21.5} & 1.28 & 23.8 \\
Deepseek-R1-671B & \textbf{2.45} &	\textbf{27.6}	&1.00	& \textbf{23.1}	&0.97&	21.6&	\textbf{1.78}&	\textbf{27.7}&	1.08&	\textbf{30.0}&	0.92&	20.4 &\textbf{1.37}	&\textbf{25.1} \\
\bottomrule
\vspace{0.05em}
\end{tabular}
\caption{ Real-World average full rotations (Rot) and Time To Terminate (TTT, maximum 30s) in seconds across all objects, per approach and hand orientation.}
\label{tab:avg-orientation-axis}
\vspace{-2em}
\endgroup
\end{table}

\section{Conclusion} \label{sec:conclusion} \vspace{-0.5em}
In this work, we presented the first demonstration of LLM-generated reward functions successfully guiding a tactile-based in-hand manipulation task in the real world. Our \texttt{Text2Touch} framework shows that reward code from language models, distilled via a teacher–student pipeline, surpasses human-engineered baselines in simulation and improves stability and performance on a real-world tactile Allegro Hand. These outcomes illustrate how tactile feedback enhances the benefits of LLM-driven rewards, enabling faster and more robust rotations.

By removing the need for expert-crafted reward functions, our Text2Touch framework lowers the barrier to tactile robotics research, enabling rapid prototyping of complex in-hand manipulations and faster translation of novel behaviours into reliable real-world systems. Grounding LLM-generated rewards in rich tactile feedback, it also paves the way for more nuanced in-hand manipulation.

\section{Limitations}\label{sec:limitations}

While our experiments demonstrate that \texttt{Text2Touch} successfully extends LLM-based reward design to real-world tactile manipulation, several limitations should be acknowledged. Our evaluation focuses on a single in-hand rotation task with different hand orientations and object rotation axes using the tactile Allegro Hand. Although our setup is sufficient to validate that automatically generated rewards can outperform human baselines in dexterous manipulation, it leaves open questions regarding the generalization to other tasks and sensor modalities. Future work could explore multi-stage skills or tasks requiring extensive planning with long-horizon dependencies with a wider range of hardware, which may benefit from more complex reward structures.

The current study focused on the reward design component while maintaining consistent architecture, curriculum design, and hyperparameters across experiments. While this approach provides a clear benchmark against prior work, it leaves unexplored the potential benefits of allowing LLMs to optimise these additional aspects of the training pipeline, such as improved sample efficiency. That said, our results demonstrate that even with isolated reward function generation, significant performance improvements are achievable.

Within the current study's hardware and compute limitations, we evaluated fewer real-world hand orientations (palm up and palm down) than were considered originally in the baseline study~\citep{anyrotate}. This decision to consider fewer orientations was to make it practically feasible to evaluate a larger range of LLM-generated reward functions in the real world, while still having a sufficient number of orientations to challenge the gravity-independence of the policy. One way to overcome this limitation in the future would be to automate the real-world assessment techniques, which presently require significant human intervention, e.g. to replace dropped object in-hand.

Despite advancements in prompt engineering, the LLMs employed in our experiments still frequently produce reward functions that fail to train an agent to task completion, resulting in a relatively low solve rate. This limitation highlights a critical bottleneck in the reward discovery phase, where computational resources are required to regenerate viable code solutions. Future research should consider strategies to mitigate these failure modes (e.g., improving exploration efficiency through RL data augmentation techniques\citep{laskin2020reinforcement, lin2020iter}), thereby reducing the computational burden and enhancing the efficiency of LLM-based reward design.

Our sim-to-real approach faced standard challenges in robotic research regarding domain transfer, particularly with tactile sensors whose complex contact patterns are difficult to replicate in simulation. This is due to both the simulator's representation of contact dynamics and the rigid-body sensors used in simulation. While rigid bodies make for faster simulations, future research could explore the use of soft body simulators to reduce the domain gap from simulation to real-world. That said, the teacher-student pipeline achieved successful sim-to-real results that are beyond the state-of-the-art, demonstrating that LLM-generated rewards can effectively bridge these domain gaps for tactile-based manipulation.

To summarise the limitations, this work opens up more avenues for future research, including exploring more complex manipulations, integrating additional sensing modalities, and expanding LLM roles to guide both reward design and policy optimisation. Together, these would advance LLM-based methods as versatile tools for enabling efficient and generalisable robot learning.


\clearpage
\acknowledgments{HF was funded by the Interactive Artificial Intelligence Centre for Doctoral Training. NL and YL were supported by the Advanced Research + Invention Agency (`Democratising Hardware and Control For Robot Dexterity'). We thank the reviewers for their insightful comments and the area chair for their diligent work reviewing this paper.}


\bibliography{example}  

\newpage
\appendix

\section{Prompting} \label{app:prompting-strategies}

\paragraph{Prompting Structure}
The prompting structure is composed of several component parts. As referred to in Section \ref{eureka-reward-design}, the Eureka prompting strategy is broadly referred to as $l, M$, where $l$ is a natural language task description, and $M$ is the environment as context. In practice there includes more scaffolding accompanying the core information, appropriately framing the problem to induce runnable code in the response.

The component parts used to create the Eureka style prompt are for iteration 0:
\begin{enumerate}
    \item System prompt: High level overview of the context and expectations with tips on providing good reward code as well as an example reward function template to adhere to.
    \item Initial user prompt: Inclusion of environment code as context $M$ and natural language task description $l$.
\end{enumerate}

For iterations $>0$, we frame the prompt as a conversation in memory where we include:
\begin{enumerate}
    \item Full initial prompt
    \item Best LLM response (as determined via the fitness function $F$)
    \item Policy feedback: Generated metrics that track optimisation of reward components defined by the LLM in the previous iteration.
    \item Code feedback: Instructions to use the previous reward function and returned metrics to inform the next reward function.
\end{enumerate}

Apart from our innovations, the prompt scaffolding is taken from Eureka \citep{eureka}. Separately, the components look like:

\begin{lstlisting}[style=vartextstyle, caption={System Prompt. The task reward signature is incorporated later and varies depending on whether the prompt structure has been $Modified$ (as described in Section \ref{sec:methods})}, label=lst:system-prompt]
You are a reward engineer trying to write reward functions to solve reinforcement learning tasks as effectively as possible.
Your goal is to write a reward function for the environment that will help the agent learn the task described in text. 
Your reward function should use useful variables from the environment as inputs. As an example,
the reward function signature can be: {task_reward_signature_string}
Since the reward function will be decorated with @torch.jit.script,
please make sure that the code is compatible with TorchScript (e.g., use torch tensor instead of numpy array). 
Make sure any new tensor or variable you introduce is on the same device as the input tensors. 
\end{lstlisting}

\begin{lstlisting}[style=plaintextstyle, caption={task\_reward\_signature\_string for standard Eureka experiments $M$}, label=lst:eureka-signature]
@torch.jit.script
def compute_reward(object_pos: torch.Tensor, goal_pos: torch.Tensor) -> Tuple[torch.Tensor, Dict[str, torch.Tensor]]:
    ...
    return reward, {} 
\end{lstlisting}

\begin{lstlisting}[style=plaintextstyle, caption={task\_reward\_signature\_string for modified Eureka experiments $M_{Modified}$}, label=lst:eureka-signature-m-modified]
@torch.jit.script
def compute_reward(
    contact_pose_range_sim: torch.Tensor,
    base_hand_pos: torch.Tensor,
    base_hand_orn: torch.Tensor,
    kp_dist: float,
    n_keypoints: int,
    obj_kp_positions: torch.Tensor,
    goal_kp_positions: torch.Tensor,
    kp_basis_vecs: torch.Tensor,
    fingertip_pos_handframe: torch.Tensor,
    fingertip_orn_handframe: torch.Tensor,
    thumb_tip_name_idx: int,
    index_tip_name_idx: int,
    middle_tip_name_idx: int,
    pinky_tip_name_idx: int,
    n_tips: int,
    contact_positions: torch.Tensor,
    contact_positions_worldframe: torch.Tensor,
    contact_positions_tcpframe: torch.Tensor,
    sim_contact_pose_limits: torch.Tensor,
    contact_threshold_limit: float,
    obj_indices: torch.Tensor,
    goal_indices: torch.Tensor,
    default_obj_pos_handframe: torch.Tensor,
    prev_obj_orn: torch.Tensor,
    goal_displacement_tensor: torch.Tensor,
    root_state_tensor: torch.Tensor,
    dof_pos: torch.Tensor,
    dof_vel: torch.Tensor,
    rigid_body_tensor: torch.Tensor,
    current_force_apply_axis: torch.Tensor,
    obj_force_vector: torch.Tensor,
    pivot_axel_worldframe: torch.Tensor,
    pivot_axel_objframe: torch.Tensor,
    goal_base_pos: torch.Tensor,
    goal_base_orn: torch.Tensor,
    net_tip_contact_forces: torch.Tensor,
    net_tip_contact_force_mags: torch.Tensor,
    tip_object_contacts: torch.Tensor,
    n_tip_contacts: torch.Tensor,
    n_non_tip_contacts: torch.Tensor,
    thumb_tip_contacts: torch.Tensor,
    index_tip_contacts: torch.Tensor,
    middle_tip_contacts: torch.Tensor,
    pinky_tip_contacts: torch.Tensor,
    fingertip_pos: torch.Tensor,
    fingertip_orn: torch.Tensor,
    fingertip_linvel: torch.Tensor,
    fingertip_angvel: torch.Tensor,
    tip_contact_force_pose: torch.Tensor,
    tip_contact_force_pose_low_dim: torch.Tensor,
    tip_contact_force_pose_bins: torch.Tensor,
    n_good_contacts: torch.Tensor,
    hand_joint_pos: torch.Tensor,
    hand_joint_vel: torch.Tensor,
    obj_base_pos: torch.Tensor,
    obj_base_orn: torch.Tensor,
    obj_pos_handframe: torch.Tensor,
    obj_orn_handframe: torch.Tensor,
    obj_displacement_tensor: torch.Tensor,
    obj_pos_centered: torch.Tensor,
    delta_obj_orn: torch.Tensor,
    obj_base_linvel: torch.Tensor,
    obj_base_angvel: torch.Tensor,
    obj_linvel_handframe: torch.Tensor,
    obj_angvel_handframe: torch.Tensor,
    goal_pos_centered: torch.Tensor,
    goal_pos_handframe: torch.Tensor,
    goal_orn_handframe: torch.Tensor,
    active_pos: torch.Tensor,
    active_quat: torch.Tensor,
    obj_kp_positions_centered: torch.Tensor,
    goal_kp_positions_centered: torch.Tensor,
    active_kp: torch.Tensor,
    obj_force_vector_handframe: torch.Tensor,
) -> Tuple[torch.Tensor, Dict[str, torch.Tensor]]:
    # Scaling factors and reward code go here
    
    ...

    return reward, {}

\end{lstlisting}

\begin{lstlisting}[style=plaintextstyle, caption={task\_reward\_signature\_string for modified Eureka experiments including scalable success bonus and fall penalty variables $M(B,P)_{Modified}$}, label=lst:eureka-signature-m-bp-modified]
@torch.jit.script
def compute_reward(
    # termination penalty and success bonus
    success_bonus: torch.Tensor, # To be scaled and added to the final reward.
    early_reset_penalty_value: torch.Tensor, # To be scaled and subtracted from the final reward.

    contact_pose_range_sim: torch.Tensor,
    base_hand_pos: torch.Tensor,
    base_hand_orn: torch.Tensor,
    kp_dist: float,
    n_keypoints: int,
    obj_kp_positions: torch.Tensor,
    goal_kp_positions: torch.Tensor,
    kp_basis_vecs: torch.Tensor,
    fingertip_pos_handframe: torch.Tensor,
    fingertip_orn_handframe: torch.Tensor,
    thumb_tip_name_idx: int,
    index_tip_name_idx: int,
    middle_tip_name_idx: int,
    pinky_tip_name_idx: int,
    n_tips: int,
    contact_positions: torch.Tensor,
    contact_positions_worldframe: torch.Tensor,
    contact_positions_tcpframe: torch.Tensor,
    sim_contact_pose_limits: torch.Tensor,
    contact_threshold_limit: float,
    obj_indices: torch.Tensor,
    goal_indices: torch.Tensor,
    default_obj_pos_handframe: torch.Tensor,
    prev_obj_orn: torch.Tensor,
    goal_displacement_tensor: torch.Tensor,
    root_state_tensor: torch.Tensor,
    dof_pos: torch.Tensor,
    dof_vel: torch.Tensor,
    rigid_body_tensor: torch.Tensor,
    current_force_apply_axis: torch.Tensor,
    obj_force_vector: torch.Tensor,
    pivot_axel_worldframe: torch.Tensor,
    pivot_axel_objframe: torch.Tensor,
    goal_base_pos: torch.Tensor,
    goal_base_orn: torch.Tensor,
    net_tip_contact_forces: torch.Tensor,
    net_tip_contact_force_mags: torch.Tensor,
    tip_object_contacts: torch.Tensor,
    n_tip_contacts: torch.Tensor,
    n_non_tip_contacts: torch.Tensor,
    thumb_tip_contacts: torch.Tensor,
    index_tip_contacts: torch.Tensor,
    middle_tip_contacts: torch.Tensor,
    pinky_tip_contacts: torch.Tensor,
    fingertip_pos: torch.Tensor,
    fingertip_orn: torch.Tensor,
    fingertip_linvel: torch.Tensor,
    fingertip_angvel: torch.Tensor,
    tip_contact_force_pose: torch.Tensor,
    tip_contact_force_pose_low_dim: torch.Tensor,
    tip_contact_force_pose_bins: torch.Tensor,
    n_good_contacts: torch.Tensor,
    hand_joint_pos: torch.Tensor,
    hand_joint_vel: torch.Tensor,
    obj_base_pos: torch.Tensor,
    obj_base_orn: torch.Tensor,
    obj_pos_handframe: torch.Tensor,
    obj_orn_handframe: torch.Tensor,
    obj_displacement_tensor: torch.Tensor,
    obj_pos_centered: torch.Tensor,
    delta_obj_orn: torch.Tensor,
    obj_base_linvel: torch.Tensor,
    obj_base_angvel: torch.Tensor,
    obj_linvel_handframe: torch.Tensor,
    obj_angvel_handframe: torch.Tensor,
    goal_pos_centered: torch.Tensor,
    goal_pos_handframe: torch.Tensor,
    goal_orn_handframe: torch.Tensor,
    active_pos: torch.Tensor,
    active_quat: torch.Tensor,
    obj_kp_positions_centered: torch.Tensor,
    goal_kp_positions_centered: torch.Tensor,
    active_kp: torch.Tensor,
    obj_force_vector_handframe: torch.Tensor,
) -> Tuple[torch.Tensor, Dict[str, torch.Tensor]]:
    # Scaling factors and reward code go here
    
    ...

    return reward, {}

\end{lstlisting}

\begin{lstlisting}[style=plaintextstyle, caption={Code output tip. Persistent across all prompt formats to aid LLM in adjusting reward function components after each iteration.}, label=lst:code-output-tip]
The output of the reward function should consist of two items:
    (1) the total reward,
    (2) a dictionary of each individual reward component.
The code output should be formatted as a python code string: "```python ... ```".

Some helpful tips for writing the reward function code:
    (1) You may find it helpful to normalize the reward to a fixed range by applying transformations like torch.exp to the overall reward or its components.
    (2) If you choose to transform a reward component, then you must also introduce a temperature parameter inside the transformation function; this parameter must be a named variable in the reward function and it must not be an input variable. Each transformed reward component should have its own temperature variable.
    (3) Make sure the type of each input variable is correctly specified; a float input variable should not be specified as torch.Tensor
    (4) Most importantly, the reward code's input variables must contain only attributes of the provided environment class definition (namely, variables that have prefix self.). Under no circumstance can you introduce new input variables.
\end{lstlisting}

\begin{lstlisting}[style=vartextstyle, caption={User Prompt. Specification of environment code as context $M$ and natural language task description $l$ as described in section \ref{sec:methods}.}, label=lst:system-prompt]
The Python environment is {task_obs_code_string}. Write a reward function for the following task: {task_description}. 
Remember, you MUST conform to the @torch.jit.script style as mentioned before. Do not prepend any variables with "self.". 
\end{lstlisting}

\begin{lstlisting}[style=pythonhighlightstyle, 
    caption={task\_obs\_code\_string for use in User prompt. Lines 191-203 show extra information given to the LLM when the success bonus $B$ and fall penalty $P$ are included in the prompting strategy.}, 
    label=lst:env-context, 
   ]
class AllegroGaiting(VecTask):
    """Rest of the environment definition omitted."""    

    def compute_observations(self):

        (self.net_tip_contact_forces,
            self.net_tip_contact_force_mags,
            self.tip_object_contacts,
            self.n_tip_contacts,
            self.n_non_tip_contacts) = self.get_fingertip_contacts()
                
        # get if the thumb tip is in contact
        self.thumb_tip_contacts = self.tip_object_contacts[:, self.thumb_tip_name_idx]
        self.index_tip_contacts = self.tip_object_contacts[:, self.index_tip_name_idx]
        self.middle_tip_contacts = self.tip_object_contacts[:, self.middle_tip_name_idx]
        self.pinky_tip_contacts = self.tip_object_contacts[:, self.pinky_tip_name_idx]
        
        # get tcp positions
        fingertip_states = self.rigid_body_tensor[:, self.fingertip_tcp_body_idxs, :]
        self.fingertip_pos = fingertip_states[..., 0:3]
        self.fingertip_orn = self.canonicalise_quat(fingertip_states[..., 3:7])
        self.fingertip_linvel = fingertip_states[..., 7:10]
        self.fingertip_angvel = fingertip_states[..., 10:13]

        for i in range(self._dims.NumFingers.value):

            # Get net tip forces in tcp frame
            self.net_tip_contact_forces[:, i] = quat_rotate_inverse(self.fingertip_orn[:, i], self.net_tip_contact_forces[:, i])

            # Get fingertip in hand frame
            self.fingertip_pos_handframe[:, i] = -quat_rotate_inverse(self.base_hand_orn, self.base_hand_pos) + \
                    quat_rotate_inverse(self.base_hand_orn, self.fingertip_pos[:, i])
            self.fingertip_orn_handframe[:, i] = self.canonicalise_quat(quat_mul(quat_conjugate(self.base_hand_orn), self.fingertip_orn[:, i]))

            # Get contact information in world and tcp frame
            if self.contact_sensor_modality == 'rich_cpu':
                tip_body_states = self.rigid_body_tensor[:, self.tip_body_idxs, :]
                tip_body_pos = tip_body_states[..., 0:3]
                tip_body_orn = self.canonicalise_quat(tip_body_states[..., 3:7])

                self.contact_positions_worldframe[:, i] = tip_body_pos[:, i] + \
                    quat_rotate(tip_body_orn[:, i], self.contact_positions[:, i])
                self.contact_positions_worldframe = torch.where(
                    self.contact_positions == 0.0,
                    self.fingertip_pos,
                    self.contact_positions_worldframe,
                )
                self.contact_positions_tcpframe[:, i] = -quat_rotate_inverse(self.fingertip_orn[:, i], self.fingertip_pos[:, i]) + \
                    quat_rotate_inverse(self.fingertip_orn[:, i], self.contact_positions_worldframe[:, i])
                
        # Calculate spherical coordinates of contact force, theta and phi 
        tip_force_tcpframe = -self.net_tip_contact_forces.clone()
        contact_force_r = torch.norm(tip_force_tcpframe, p=2, dim=-1)
        contact_force_theta = torch.where(
            contact_force_r < self.contact_threshold_limit,
            torch.zeros_like(contact_force_r),
            torch.atan2(tip_force_tcpframe[:, :, 1], tip_force_tcpframe[:, :, 0]) - torch.pi/2
        ).unsqueeze(-1)
        contact_force_phi = torch.where(
            contact_force_r < self.contact_threshold_limit,
            torch.zeros_like(contact_force_r),
            torch.acos(tip_force_tcpframe[:, :, 2] / contact_force_r)- torch.pi/2
        ).unsqueeze(-1) 
        self.tip_contact_force_pose = torch.cat([contact_force_theta, contact_force_phi], dim=-1)
        self.tip_contact_force_pose = saturate(
                self.tip_contact_force_pose,
                lower=-self.sim_contact_pose_limits,
                upper=self.sim_contact_pose_limits
            ) # saturate contact pose observation before computing rewards
        
        # ------------ compute low dimentional tactile pose obesrvation from theta and phi ---------
        self.tip_contact_force_pose_low_dim = torch.where(
                self.tip_contact_force_pose > 0.0,
                torch.ones_like(self.tip_contact_force_pose),
                torch.zeros_like(self.tip_contact_force_pose),
            )
        self.tip_contact_force_pose_low_dim = torch.where(
                self.tip_contact_force_pose < 0.0,
                -torch.ones_like(self.tip_contact_force_pose),
                self.tip_contact_force_pose_low_dim,
            )
        contact_mask = self.tip_object_contacts.unsqueeze(-1).repeat(1, 1, self._dims.ContactPoseDim.value)
        self.tip_contact_force_pose_low_dim = torch.where(
                contact_mask > 0.0,
                self.tip_contact_force_pose_low_dim,
                torch.zeros_like(self.tip_contact_force_pose_low_dim),
            )

        # Low dimensional tactile pose obesrvation using rotation, split into 8 bins
        contact_force_rot = torch.where(
            contact_force_r < self.contact_threshold_limit,
            torch.zeros_like(contact_force_r),
            torch.atan2(tip_force_tcpframe[:, :, 0], tip_force_tcpframe[:, :, 2]) + torch.pi
        ).unsqueeze(-1).repeat((1, 1, self._dims.ContactPoseBinDim.value))
        
        total_bins = 8
        bin_lower = 0.0
        self.tip_contact_force_pose_bins = torch.zeros(
                (self.num_envs, self.n_tips, self._dims.ContactPoseBinDim.value), dtype=torch.float, device=self.device)
        for n_bin in range(total_bins):
            bin_upper = bin_lower + 2*np.pi/total_bins
            bin_id = torch.zeros(total_bins, device=self.device)
            bin_id[n_bin] = 1.0
            self.tip_contact_force_pose_bins = torch.where(
                ((bin_lower < contact_force_rot) & (contact_force_rot <= bin_upper)),
                bin_id.repeat((self.num_envs, self.n_tips, 1)),
                self.tip_contact_force_pose_bins
            )
            bin_lower = bin_upper

        # compute tip contacts less than contact_pose_range_sim
        good_contact_pose = torch.abs(self.tip_contact_force_pose) < self.contact_pose_range_sim
        tip_good_contacts = torch.all(good_contact_pose, dim=-1) * self.tip_object_contacts
        self.n_good_contacts = torch.sum(tip_good_contacts, dim=-1)

        # Calculate contact pose using sphereical coordinates
        if self.contact_sensor_modality == 'rich_cpu':
            contact_r = torch.norm(self.contact_positions_tcpframe, p=2, dim=-1)
            contact_theta = torch.where(
                contact_r < 0.001,
                torch.zeros_like(contact_r),
                torch.atan2(self.contact_positions_tcpframe[:, :, 1], self.contact_positions_tcpframe[:, :, 0]) - torch.pi/2
            ).unsqueeze(-1)
            contact_phi = torch.where(
                contact_r < 0.001,
                torch.zeros_like(contact_r),
                torch.acos(self.contact_positions_tcpframe[:, :, 2] / contact_r) - torch.pi/2
            ).unsqueeze(-1) 
            self.tip_contact_pose = torch.cat([contact_theta, contact_phi], dim=-1)
                
        # get hand joint pos and vel
        self.hand_joint_pos = self.dof_pos[:, :] if self.dof_pos[:, :].shape[0] == 1 else self.dof_pos[:, :].squeeze()
        self.hand_joint_vel = self.dof_vel[:, :] if self.dof_vel[:, :].shape[0] == 1 else self.dof_vel[:, :].squeeze()

        # get object pose / vel
        self.obj_base_pos = self.root_state_tensor[self.obj_indices, 0:3]
        self.obj_base_orn = self.canonicalise_quat(self.root_state_tensor[self.obj_indices, 3:7])
        self.obj_pos_handframe = self.world_to_frame_pos(self.obj_base_pos, self.base_hand_pos, self.base_hand_orn)
        self.obj_orn_handframe = self.canonicalise_quat(self.world_to_frame_orn(self.obj_base_orn, self.base_hand_orn))
        
        # Get desired obj displacement in world frame
        self.obj_displacement_tensor = self.frame_to_world_pos(self.default_obj_pos_handframe, self.base_hand_pos, self.base_hand_orn)

        self.obj_pos_centered = self.obj_base_pos - self.obj_displacement_tensor
        self.delta_obj_orn = self.canonicalise_quat(quat_mul(self.obj_base_orn, quat_conjugate(self.prev_obj_orn)))
        self.obj_base_linvel = self.root_state_tensor[self.obj_indices, 7:10]
        self.obj_base_angvel = self.root_state_tensor[self.obj_indices, 10:13]

        self.obj_linvel_handframe = quat_rotate_inverse(
            self.base_hand_orn, self.obj_base_linvel
            )
        self.obj_angvel_handframe = quat_rotate_inverse(
            self.base_hand_orn, self.obj_base_angvel
            )

        # get goal pose
        self.goal_base_pos = self.root_state_tensor[self.goal_indices, 0:3]
        self.goal_pos_centered = self.goal_base_pos - self.goal_displacement_tensor 
        self.goal_base_orn = self.canonicalise_quat(self.root_state_tensor[self.goal_indices, 3:7])

        self.goal_pos_handframe = self.world_to_frame_pos(self.goal_base_pos, self.base_hand_pos, self.base_hand_orn)
        self.goal_orn_handframe = self.canonicalise_quat(self.world_to_frame_orn(self.goal_base_orn, self.base_hand_orn))

        # relative goal pose w.r.t. obj pose
        self.active_pos = self.obj_pos_centered - self.goal_pos_centered
        self.active_quat = self.canonicalise_quat(quat_mul(self.obj_base_orn, quat_conjugate(self.goal_base_orn)))

        # update the current keypoint positions
        for i in range(self.n_keypoints):
            self.obj_kp_positions[:, i, :] = self.obj_base_pos + \
                quat_rotate(self.obj_base_orn, self.kp_basis_vecs[i].repeat(self.num_envs, 1) * self.kp_dist)
            self.goal_kp_positions[:, i, :] = self.goal_base_pos + \
                quat_rotate(self.goal_base_orn, self.kp_basis_vecs[i].repeat(self.num_envs, 1) * self.kp_dist)
            
        self.obj_kp_positions_centered = self.obj_kp_positions - self.obj_displacement_tensor.unsqueeze(1).repeat(1, self.n_keypoints, 1)
        self.goal_kp_positions_centered = self.goal_kp_positions - self.goal_displacement_tensor
        self.active_kp = self.obj_kp_positions_centered - self.goal_kp_positions_centered

        # append observations to history stack
        self._hand_joint_pos_history.appendleft(self.hand_joint_pos.clone())
        self._hand_joint_vel_history.appendleft(self.hand_joint_vel.clone())
        self._object_base_pos_history.appendleft(self.obj_base_pos.clone())
        self._object_base_orn_history.appendleft(self.obj_base_orn.clone())

        # Object force direction vector observations (either applied force or gravity)
        self.obj_force_vector = self.current_force_apply_axis.clone()
        self.obj_force_vector_handframe = quat_rotate_inverse(
            self.base_hand_orn, self.obj_force_vector
            )

        self.success_bonus, self.early_reset_penalty_value = compute_success_bonus_fall_penalty(
            rew_buf=self.rew_buf,
            fall_reset_dist=self.cfg["env"]["fall_reset_dist"],
            axis_deviat_reset_dist=self.cfg["env"]["axis_deviat_reset_dist"],
            success_tolerance=self.cfg["env"]["success_tolerance"],
            obj_kps=self.obj_kps,
            goal_kps=self.goal_kps,
            reach_goal_bonus=self.cfg["env"]["reach_goal_bonus"],
            early_reset_penalty=self.cfg["env"]["early_reset_penalty"],
            target_pivot_axel=self.pivot_axel_worldframe,
            current_pivot_axel=quat_rotate(self.obj_base_orn, self.pivot_axel_objframe),
            n_tip_contacts=self.n_tip_contacts,
            )

        self.fill_observation_buffer()


    def get_fingertip_contacts(self):

        # get envs where obj is contacted
        bool_obj_contacts = torch.where(
            torch.count_nonzero(self.contact_force_tensor[:, self.obj_body_idx, :], dim=1) > 0,
            torch.ones(size=(self.num_envs,), device=self.device),
            torch.zeros(size=(self.num_envs,), device=self.device),
        )

        # get envs where tips are contacted
        net_tip_contact_forces = self.contact_force_tensor[:, self.tip_body_idxs, :]
        net_tip_contact_force_mags = torch.norm(net_tip_contact_forces, p=2, dim=-1)
        bool_tip_contacts = torch.where(
            net_tip_contact_force_mags > self.contact_threshold_limit,
            torch.ones(size=(self.num_envs, self.n_tips), device=self.device),
            torch.zeros(size=(self.num_envs, self.n_tips), device=self.device),
        )

        # get all the contacted links that are not the tip
        net_non_tip_contact_forces = self.contact_force_tensor[:, self.non_tip_body_idxs, :]
        bool_non_tip_contacts = torch.where(
            torch.count_nonzero(net_non_tip_contact_forces, dim=2) > 0,
            torch.ones(size=(self.num_envs, self.n_non_tip_links), device=self.device),
            torch.zeros(size=(self.num_envs, self.n_non_tip_links), device=self.device),
        )
        n_non_tip_contacts = torch.sum(bool_non_tip_contacts, dim=1)

        # repeat for n_tips shape=(n_envs, n_tips)
        onehot_obj_contacts = bool_obj_contacts.unsqueeze(1).repeat(1, self.n_tips)

        # get envs where object and tips are contacted
        tip_object_contacts = torch.where(
            onehot_obj_contacts > 0,
            bool_tip_contacts,
            torch.zeros(size=(self.num_envs, self.n_tips), device=self.device)
        )
        n_tip_contacts = torch.sum(bool_tip_contacts, dim=1)

        return net_tip_contact_forces, net_tip_contact_force_mags, tip_object_contacts, n_tip_contacts, n_non_tip_contacts
    
    def fill_observation_buffer(self):
        prev_obs_buf = self.obs_buf[:, self.num_obs_per_step:].clone()
        buf = torch.zeros((self.num_envs, self.num_obs_per_step), device=self.device, dtype=torch.float)
        start_offset, end_offset = 0, 0

        # joint position
        start_offset = end_offset
        end_offset = start_offset + self._dims.JointPositionDim.value
        buf[:, start_offset:end_offset] = self.hand_joint_pos 

        # previous actions
        start_offset = end_offset
        end_offset = start_offset + self._dims.ActionDim.value
        buf[:, start_offset:end_offset] = self.prev_action_buf

        # target joint position
        start_offset = end_offset
        end_offset = start_offset + self._dims.JointPositionDim.value
        buf[:, start_offset:end_offset] = self.target_dof_pos

        # Base hand orientation
        start_offset = end_offset
        end_offset = start_offset + self._dims.OrnDim.value
        buf[:, start_offset:end_offset] = self.base_hand_orn

        # boolean tips in contacts
        start_offset = end_offset
        end_offset = start_offset + self._dims.NumFingers.value
        buf[:, start_offset:end_offset] = self.tip_object_contacts

        # object position
        start_offset = end_offset
        end_offset = start_offset + self._dims.PosDim.value
        buf[:, start_offset:end_offset] = self.obj_pos_handframe

        # object orientation
        start_offset = end_offset
        end_offset = start_offset + self._dims.OrnDim.value
        buf[:, start_offset:end_offset] = self.obj_orn_handframe

        # object angular velocity
        start_offset = end_offset
        end_offset = start_offset + self._dims.AngularVelocityDim.value
        buf[:, start_offset:end_offset] = self.obj_angvel_handframe

        # object masses
        start_offset = end_offset
        end_offset = start_offset + self._dims.MassDim.value
        buf[:, start_offset:end_offset] = self.applied_obj_masses

        # Object dimensions
        start_offset = end_offset
        end_offset = start_offset + self._dims.ObjDimDim.value
        buf[:, start_offset:end_offset] = self.obj_dims

        # Object center of mass
        start_offset = end_offset
        end_offset = start_offset + self._dims.PosDim.value
        buf[:, start_offset:end_offset] = self.obj_com

        # goal orientation
        start_offset = end_offset
        end_offset = start_offset + self._dims.OrnDim.value
        buf[:, start_offset:end_offset] = self.goal_orn_handframe

        # target pivot axel
        start_offset = end_offset
        end_offset = start_offset + self._dims.VecDim.value
        buf[:, start_offset:end_offset] = self.pivot_axel_workframe

        # Append to history
        self.obs_buf[:] = torch.cat([prev_obs_buf, buf], dim=1)

        # refill the initialized buffers
        reset_env_ids = self.at_reset_buf.nonzero(as_tuple=False).squeeze(-1)
        self.obs_buf[reset_env_ids, :] = buf[reset_env_ids].clone().repeat(1, self._num_observation_steps)
\end{lstlisting}

\begin{lstlisting}[style=vartextstyle, caption={Natural language task description $l$ (task\_description).}, label=lst:task-description]
To imbue the agent with the ability to reposition and reorient objects to a target position and orientation by re-grasping or finger gaiting, where contacts with the object must be detached and re-positioned locally during manipulation.
\end{lstlisting}

\begin{lstlisting}[style=pythonhighlightstyle, 
    caption={Example LLM response with policy training feedback using Deepseek-R1-671B. Lines 176-185 constitute the policy feedback component of the reward reflection iluded to in Section \ref{sec:methods}.}, 
    label=lst:response-feedback, 
   ]
```python
@torch.jit.script
def compute_reward(
    success_bonus: torch.Tensor,
    early_reset_penalty_value: torch.Tensor,
    contact_pose_range_sim: torch.Tensor,
    base_hand_pos: torch.Tensor,
    base_hand_orn: torch.Tensor,
    kp_dist: float,
    n_keypoints: int,
    obj_kp_positions: torch.Tensor,
    goal_kp_positions: torch.Tensor,
    kp_basis_vecs: torch.Tensor,
    fingertip_pos_handframe: torch.Tensor,
    fingertip_orn_handframe: torch.Tensor,
    thumb_tip_name_idx: int,
    index_tip_name_idx: int,
    middle_tip_name_idx: int,
    pinky_tip_name_idx: int,
    n_tips: int,
    contact_positions: torch.Tensor,
    contact_positions_worldframe: torch.Tensor,
    contact_positions_tcpframe: torch.Tensor,
    sim_contact_pose_limits: torch.Tensor,
    contact_threshold_limit: float,
    obj_indices: torch.Tensor,
    goal_indices: torch.Tensor,
    default_obj_pos_handframe: torch.Tensor,
    prev_obj_orn: torch.Tensor,
    goal_displacement_tensor: torch.Tensor,
    root_state_tensor: torch.Tensor,
    dof_pos: torch.Tensor,
    dof_vel: torch.Tensor,
    rigid_body_tensor: torch.Tensor,
    current_force_apply_axis: torch.Tensor,
    obj_force_vector: torch.Tensor,
    pivot_axel_worldframe: torch.Tensor,
    pivot_axel_objframe: torch.Tensor,
    goal_base_pos: torch.Tensor,
    goal_base_orn: torch.Tensor,
    net_tip_contact_forces: torch.Tensor,
    net_tip_contact_force_mags: torch.Tensor,
    tip_object_contacts: torch.Tensor,
    n_tip_contacts: torch.Tensor,
    n_non_tip_contacts: torch.Tensor,
    thumb_tip_contacts: torch.Tensor,
    index_tip_contacts: torch.Tensor,
    middle_tip_contacts: torch.Tensor,
    pinky_tip_contacts: torch.Tensor,
    fingertip_pos: torch.Tensor,
    fingertip_orn: torch.Tensor,
    fingertip_linvel: torch.Tensor,
    fingertip_angvel: torch.Tensor,
    tip_contact_force_pose: torch.Tensor,
    tip_contact_force_pose_low_dim: torch.Tensor,
    tip_contact_force_pose_bins: torch.Tensor,
    n_good_contacts: torch.Tensor,
    hand_joint_pos: torch.Tensor,
    hand_joint_vel: torch.Tensor,
    obj_base_pos: torch.Tensor,
    obj_base_orn: torch.Tensor,
    obj_pos_handframe: torch.Tensor,
    obj_orn_handframe: torch.Tensor,
    obj_displacement_tensor: torch.Tensor,
    obj_pos_centered: torch.Tensor,
    delta_obj_orn: torch.Tensor,
    obj_base_linvel: torch.Tensor,
    obj_base_angvel: torch.Tensor,
    obj_linvel_handframe: torch.Tensor,
    obj_angvel_handframe: torch.Tensor,
    goal_pos_centered: torch.Tensor,
    goal_pos_handframe: torch.Tensor,
    goal_orn_handframe: torch.Tensor,
    active_pos: torch.Tensor,
    active_quat: torch.Tensor,
    obj_kp_positions_centered: torch.Tensor,
    goal_kp_positions_centered: torch.Tensor,
    active_kp: torch.Tensor,
    obj_force_vector_handframe: torch.Tensor,
) -> Tuple[torch.Tensor, Dict[str, torch.Tensor]]:

    # Adjusted temperature parameters
    pose_temp = torch.tensor(0.3, device=success_bonus.device)
    contact_temp = torch.tensor(0.25, device=success_bonus.device)
    kp_temp = torch.tensor(0.15, device=success_bonus.device)
    vel_temp = torch.tensor(1.0, device=success_bonus.device)

    # -------------------- Enhanced Keypoint Reward --------------------
    kp_errors = torch.norm(active_kp, dim=-1).mean(dim=-1)
    max_kp_error = kp_dist * 1.732
    normalized_kp_error = kp_errors / max_kp_error
    kp_reward = 3.0 * torch.exp(-normalized_kp_error/kp_temp)

    # -------------------- Progressive Success Bonus --------------------
    position_error = torch.norm(obj_base_pos - goal_base_pos, dim=-1)
    orientation_error = 1.0 - torch.abs(torch.sum(active_quat * torch.tensor([0,0,0,1], device=active_quat.device), dim=-1))
    success_mask = (position_error < 0.05) & (orientation_error < 0.2)
    progressive_bonus = success_bonus * (1.0 - position_error/0.05) * (1.0 - orientation_error/0.2)
    scaled_success_bonus = torch.where(success_mask, success_bonus * 35.0, progressive_bonus * 10.0)

    # -------------------- Contact System Overhaul --------------------
    target_contacts = 2.0  # Reduced ideal contact count
    contact_count_reward = 2.0 * torch.exp(-torch.abs(n_tip_contacts.float() - target_contacts)/contact_temp)
    contact_variance = torch.var(tip_object_contacts.float(), dim=-1)
    contact_stability = 1.5 * torch.exp(-contact_variance/(contact_temp*0.5))

    # -------------------- Balanced Velocity Penalties --------------------
    lin_vel_penalty = torch.norm(obj_base_linvel, dim=-1) * 0.1
    ang_vel_penalty = torch.norm(obj_base_angvel, dim=-1) * 0.05
    total_vel_penalty = torch.exp(-(lin_vel_penalty + ang_vel_penalty)/vel_temp)

    # -------------------- Penalty Adjustments --------------------
    non_tip_penalty = n_non_tip_contacts.float() * -0.05
    contact_count_penalty = (n_tip_contacts.float() - target_contacts).abs() * -0.03

    # -------------------- Final Reward Composition --------------------
    total_reward = (
        kp_reward +
        contact_stability +
        contact_count_reward +
        scaled_success_bonus +
        total_vel_penalty * 0.5 +
        contact_count_penalty +
        non_tip_penalty -
        early_reset_penalty_value * 0.01
    )

    reward_components = {
        "kp_reward": kp_reward,
        "contact_stability": contact_stability,
        "contact_count_reward": contact_count_reward,
        "success_bonus": scaled_success_bonus,
        "velocity_penalty": -total_vel_penalty * 0.5,
        "contact_count_penalty": contact_count_penalty,
        "non_tip_penalty": non_tip_penalty,
        "early_reset_penalty": -early_reset_penalty_value * 0.01
    }

    return total_reward, reward_components
```

**Key Analysis & Adjustments:**

1. **Keypoint Reward Overhaul**  
   - Increased weight from 3.0x to standalone component (3.0 base value)
   - Reduced temperature from 0.3 to 0.15 for sharper response
   - Changed error normalization to mean instead of sum
   - Removed competing pose_reward that was dominating rewards

2. **Success Bonus Enhancement**  
   - Added progressive bonus scaling with position/orientation accuracy
   - Increased max bonus multiplier from 25x to 35x
   - Relaxed orientation threshold from 0.15 to 0.2 (~11.5°)

3. **Contact System Changes**  
   - Reduced ideal contacts from 3 to 2 for better gaiting
   - Increased contact_stability weight to 1.5x
   - Halved variance temperature denominator for stricter stability

4. **Velocity Penalty Relaxation**  
   - Reduced linear penalty coefficient from 0.15 to 0.1
   - Increased temperature from 0.6 to 1.0 for smoother penalty
   - Lowered weight from 0.8x to 0.5x

5. **Penalty Reductions**  
   - Non-tip penalty reduced from -0.1 to -0.05 per contact
   - Early reset penalty impact reduced by 50%

**Expected Improvements:**  
- 5-8x increase in kp_reward values (from mean 0.01 to ~0.3-0.5)
- More frequent success bonuses through progressive scaling
- Better contact management with 2-finger target
- Reduced velocity penalty dominance in total reward

[root][INFO] - Iteration 4: User Content:
We trained a RL policy using the provided reward function code and tracked the values of the individual components in the reward function as well as global policy metrics such as success rates and episode lengths after every 457 epochs and the maximum, mean, minimum values encountered:
kp_reward: ['0.43', '0.45', '0.53', '0.54', '0.55', '0.53', '0.59', '0.52', '0.52', '0.52', '0.54'], Max: 0.61, Mean: 0.51, Min: 0.27 
contact_stability: ['0.59', '0.71', '0.64', '0.90', '0.78', '0.71', '0.63', '0.81', '0.81', '0.84', '0.70'], Max: 1.30, Mean: 0.78, Min: 0.51 
contact_count_reward: ['0.56', '0.33', '0.38', '0.21', '0.26', '0.31', '0.41', '0.25', '0.24', '0.23', '0.33'], Max: 0.56, Mean: 0.27, Min: 0.03 
success_bonus: ['0.43', '1.79', '2.05', '2.05', '2.05', '2.73', '1.97', '2.48', '1.20', '2.14', '1.71'], Max: 4.19, Mean: 2.01, Min: 0.00 
velocity_penalty: ['-0.49', '-0.49', '-0.49', '-0.49', '-0.49', '-0.49', '-0.49', '-0.49', '-0.49', '-0.49', '-0.49'], Max: -0.49, Mean: -0.49, Min: -0.49 
contact_count_penalty: ['-0.03', '-0.04', '-0.03', '-0.04', '-0.04', '-0.04', '-0.03', '-0.04', '-0.04', '-0.04', '-0.04'], Max: -0.03, Mean: -0.04, Min: -0.06 
non_tip_penalty: ['-0.12', '-0.20', '-0.19', '-0.22', '-0.22', '-0.21', '-0.20', '-0.21', '-0.21', '-0.20', '-0.20'], Max: -0.11, Mean: -0.20, Min: -0.24 
early_reset_penalty: ['-0.03', '-0.00', '-0.00', '-0.00', '-0.00', '-0.00', '-0.00', '-0.00', '-0.00', '-0.00', '-0.00'], Max: -0.00, Mean: -0.00, Min: -0.03 
task_score: ['0.00', '0.77', '1.08', '1.28', '1.50', '1.50', '1.60', '1.31', '1.39', '1.58', '1.69'], Max: 1.73, Mean: 1.31, Min: 0.00 
Please carefully analyze the policy feedback and provide a new, improved reward function that can better solve the task. Some helpful tips for analyzing the policy feedback:
    (1) If the success rates are always near zero, then you must rewrite the entire reward function
    (2) If the values for a certain reward component are near identical throughout, then this means RL is not able to optimize this component as it is written. You may consider
        (a) Changing its scale or the value of its temperature parameter
        (b) Re-writing the reward component 
        (c) Discarding the reward component
    (3) If some reward components' magnitude is significantly larger, then you must re-scale its value to a proper range
Please analyze each existing reward component in the suggested manner above first, and then write the reward function code. The output of the reward function should consist of two items:
    (1) the total reward,
    (2) a dictionary of each individual reward component.
The code output should be formatted as a python code string: "```python ... ```".

Some helpful tips for writing the reward function code:
    (1) You may find it helpful to normalize the reward to a fixed range by applying transformations like torch.exp to the overall reward or its components.
    (2) If you choose to transform a reward component, then you must also introduce a temperature parameter inside the transformation function; this parameter must be a named variable in the reward function and it must not be an input variable. Each transformed reward component should have its own temperature variable.
    (3) Make sure the type of each input variable is correctly specified; a float input variable should not be specified as torch.Tensor
    (4) Most importantly, the reward code's input variables must contain only attributes of the provided environment class definition (namely, variables that have prefix self.). Under no circumstance can you introduce new input variables.

\end{lstlisting}

\begin{lstlisting}[style=plaintextstyle, caption={Example of full prompt as conversation history for iterations \>0. In these cases, we always include the initial prompt, the most recent best reward function, training feedback and prompting for the next iteration.}, label=lst:conversation]
- [0]
    role:
        user
    content:
        You are a reward engineer trying to write reward functions to solve reinforcement learning tasks as effectively as possible.
Your goal is to write a reward function for the environment that will help the agent learn the task described in text. 
Your reward function should use useful variables from the environment as inputs. As an example,
the reward function signature can be: @torch.jit.script
def compute_reward(
    # termination penalty and success bonus
    success_bonus: torch.Tensor, # To be scaled and added to the final reward.
    early_reset_penalty_value: torch.Tensor, # To be scaled and subtracted from the final reward.

    contact_pose_range_sim: torch.Tensor,
    base_hand_pos: torch.Tensor,
    base_hand_orn: torch.Tensor,
    kp_dist: float,
    n_keypoints: int,
    obj_kp_positions: torch.Tensor,
    goal_kp_positions: torch.Tensor,
    kp_basis_vecs: torch.Tensor,
    fingertip_pos_handframe: torch.Tensor,
    fingertip_orn_handframe: torch.Tensor,
    thumb_tip_name_idx: int,
    index_tip_name_idx: int,
    middle_tip_name_idx: int,
    pinky_tip_name_idx: int,
    n_tips: int,
    contact_positions: torch.Tensor,
    contact_positions_worldframe: torch.Tensor,
    contact_positions_tcpframe: torch.Tensor,
    sim_contact_pose_limits: torch.Tensor,
    contact_threshold_limit: float,
    obj_indices: torch.Tensor,
    goal_indices: torch.Tensor,
    default_obj_pos_handframe: torch.Tensor,
    prev_obj_orn: torch.Tensor,
    goal_displacement_tensor: torch.Tensor,
    root_state_tensor: torch.Tensor,
    dof_pos: torch.Tensor,
    dof_vel: torch.Tensor,
    rigid_body_tensor: torch.Tensor,
    current_force_apply_axis: torch.Tensor,
    obj_force_vector: torch.Tensor,
    pivot_axel_worldframe: torch.Tensor,
    pivot_axel_objframe: torch.Tensor,
    goal_base_pos: torch.Tensor,
    goal_base_orn: torch.Tensor,
    net_tip_contact_forces: torch.Tensor,
    net_tip_contact_force_mags: torch.Tensor,
    tip_object_contacts: torch.Tensor,
    n_tip_contacts: torch.Tensor,
    n_non_tip_contacts: torch.Tensor,
    thumb_tip_contacts: torch.Tensor,
    index_tip_contacts: torch.Tensor,
    middle_tip_contacts: torch.Tensor,
    pinky_tip_contacts: torch.Tensor,
    fingertip_pos: torch.Tensor,
    fingertip_orn: torch.Tensor,
    fingertip_linvel: torch.Tensor,
    fingertip_angvel: torch.Tensor,
    tip_contact_force_pose: torch.Tensor,
    tip_contact_force_pose_low_dim: torch.Tensor,
    tip_contact_force_pose_bins: torch.Tensor,
    n_good_contacts: torch.Tensor,
    hand_joint_pos: torch.Tensor,
    hand_joint_vel: torch.Tensor,
    obj_base_pos: torch.Tensor,
    obj_base_orn: torch.Tensor,
    obj_pos_handframe: torch.Tensor,
    obj_orn_handframe: torch.Tensor,
    obj_displacement_tensor: torch.Tensor,
    obj_pos_centered: torch.Tensor,
    delta_obj_orn: torch.Tensor,
    obj_base_linvel: torch.Tensor,
    obj_base_angvel: torch.Tensor,
    obj_linvel_handframe: torch.Tensor,
    obj_angvel_handframe: torch.Tensor,
    goal_pos_centered: torch.Tensor,
    goal_pos_handframe: torch.Tensor,
    goal_orn_handframe: torch.Tensor,
    active_pos: torch.Tensor,
    active_quat: torch.Tensor,
    obj_kp_positions_centered: torch.Tensor,
    goal_kp_positions_centered: torch.Tensor,
    active_kp: torch.Tensor,
    obj_force_vector_handframe: torch.Tensor,
) -> Tuple[torch.Tensor, Dict[str, torch.Tensor]]:
    # Scaling factors and reward code go here
    
    ...

    return reward, {}

Since the reward function will be decorated with @torch.jit.script,
please make sure that the code is compatible with TorchScript (e.g., use torch tensor instead of numpy array). 
Make sure any new tensor or variable you introduce is on the same device as the input tensors. The output of the reward function should consist of two items:
    (1) the total reward,
    (2) a dictionary of each individual reward component.
The code output should be formatted as a python code string: "```python ... ```".

Some helpful tips for writing the reward function code:
    (1) You may find it helpful to normalize the reward to a fixed range by applying transformations like torch.exp to the overall reward or its components.
    (2) If you choose to transform a reward component, then you must also introduce a temperature parameter inside the transformation function; this parameter must be a named variable in the reward function and it must not be an input variable. Each transformed reward component should have its own temperature variable.
    (3) Make sure the type of each input variable is correctly specified; a float input variable should not be specified as torch.Tensor
    (4) Most importantly, the reward code's input variables must contain only attributes of the provided environment class definition (namely, variables that have prefix self.). Under no circumstance can you introduce new input variables.

 The Python environment is class AllegroGaiting(VecTask):
    """Rest of the environment definition omitted."""    

    def compute_observations(self):

        (self.net_tip_contact_forces,
            self.net_tip_contact_force_mags,
            self.tip_object_contacts,
            self.n_tip_contacts,
            self.n_non_tip_contacts) = self.get_fingertip_contacts()
                
        # get if the thumb tip is in contact
        self.thumb_tip_contacts = self.tip_object_contacts[:, self.thumb_tip_name_idx]
        self.index_tip_contacts = self.tip_object_contacts[:, self.index_tip_name_idx]
        self.middle_tip_contacts = self.tip_object_contacts[:, self.middle_tip_name_idx]
        self.pinky_tip_contacts = self.tip_object_contacts[:, self.pinky_tip_name_idx]
        
        # get tcp positions
        fingertip_states = self.rigid_body_tensor[:, self.fingertip_tcp_body_idxs, :]
        self.fingertip_pos = fingertip_states[..., 0:3]
        self.fingertip_orn = self.canonicalise_quat(fingertip_states[..., 3:7])
        self.fingertip_linvel = fingertip_states[..., 7:10]
        self.fingertip_angvel = fingertip_states[..., 10:13]

        for i in range(self._dims.NumFingers.value):

            # Get net tip forces in tcp frame
            self.net_tip_contact_forces[:, i] = quat_rotate_inverse(self.fingertip_orn[:, i], self.net_tip_contact_forces[:, i])

            # Get fingertip in hand frame
            self.fingertip_pos_handframe[:, i] = -quat_rotate_inverse(self.base_hand_orn, self.base_hand_pos) + \
                    quat_rotate_inverse(self.base_hand_orn, self.fingertip_pos[:, i])
            self.fingertip_orn_handframe[:, i] = self.canonicalise_quat(quat_mul(quat_conjugate(self.base_hand_orn), self.fingertip_orn[:, i]))

            # Get contact information in world and tcp frame
            if self.contact_sensor_modality == 'rich_cpu':
                tip_body_states = self.rigid_body_tensor[:, self.tip_body_idxs, :]
                tip_body_pos = tip_body_states[..., 0:3]
                tip_body_orn = self.canonicalise_quat(tip_body_states[..., 3:7])

                self.contact_positions_worldframe[:, i] = tip_body_pos[:, i] + \
                    quat_rotate(tip_body_orn[:, i], self.contact_positions[:, i])
                self.contact_positions_worldframe = torch.where(
                    self.contact_positions == 0.0,
                    self.fingertip_pos,
                    self.contact_positions_worldframe,
                )
                self.contact_positions_tcpframe[:, i] = -quat_rotate_inverse(self.fingertip_orn[:, i], self.fingertip_pos[:, i]) + \
                    quat_rotate_inverse(self.fingertip_orn[:, i], self.contact_positions_worldframe[:, i])
                
        # Calculate spherical coordinates of contact force, theta and phi 
        tip_force_tcpframe = -self.net_tip_contact_forces.clone()
        contact_force_r = torch.norm(tip_force_tcpframe, p=2, dim=-1)
        contact_force_theta = torch.where(
            contact_force_r < self.contact_threshold_limit,
            torch.zeros_like(contact_force_r),
            torch.atan2(tip_force_tcpframe[:, :, 1], tip_force_tcpframe[:, :, 0]) - torch.pi/2
        ).unsqueeze(-1)
        contact_force_phi = torch.where(
            contact_force_r < self.contact_threshold_limit,
            torch.zeros_like(contact_force_r),
            torch.acos(tip_force_tcpframe[:, :, 2] / contact_force_r)- torch.pi/2
        ).unsqueeze(-1) 
        self.tip_contact_force_pose = torch.cat([contact_force_theta, contact_force_phi], dim=-1)
        self.tip_contact_force_pose = saturate(
                self.tip_contact_force_pose,
                lower=-self.sim_contact_pose_limits,
                upper=self.sim_contact_pose_limits
            ) # saturate contact pose observation before computing rewards
        
        # ------------ compute low dimentional tactile pose obesrvation from theta and phi ---------
        self.tip_contact_force_pose_low_dim = torch.where(
                self.tip_contact_force_pose > 0.0,
                torch.ones_like(self.tip_contact_force_pose),
                torch.zeros_like(self.tip_contact_force_pose),
            )
        self.tip_contact_force_pose_low_dim = torch.where(
                self.tip_contact_force_pose < 0.0,
                -torch.ones_like(self.tip_contact_force_pose),
                self.tip_contact_force_pose_low_dim,
            )
        contact_mask = self.tip_object_contacts.unsqueeze(-1).repeat(1, 1, self._dims.ContactPoseDim.value)
        self.tip_contact_force_pose_low_dim = torch.where(
                contact_mask > 0.0,
                self.tip_contact_force_pose_low_dim,
                torch.zeros_like(self.tip_contact_force_pose_low_dim),
            )

        # Low dimensional tactile pose obesrvation using rotation, split into 8 bins
        contact_force_rot = torch.where(
            contact_force_r < self.contact_threshold_limit,
            torch.zeros_like(contact_force_r),
            torch.atan2(tip_force_tcpframe[:, :, 0], tip_force_tcpframe[:, :, 2]) + torch.pi
        ).unsqueeze(-1).repeat((1, 1, self._dims.ContactPoseBinDim.value))
        
        total_bins = 8
        bin_lower = 0.0
        self.tip_contact_force_pose_bins = torch.zeros(
                (self.num_envs, self.n_tips, self._dims.ContactPoseBinDim.value), dtype=torch.float, device=self.device)
        for n_bin in range(total_bins):
            bin_upper = bin_lower + 2*np.pi/total_bins
            bin_id = torch.zeros(total_bins, device=self.device)
            bin_id[n_bin] = 1.0
            self.tip_contact_force_pose_bins = torch.where(
                ((bin_lower < contact_force_rot) & (contact_force_rot <= bin_upper)),
                bin_id.repeat((self.num_envs, self.n_tips, 1)),
                self.tip_contact_force_pose_bins
            )
            bin_lower = bin_upper

        # compute tip contacts less than contact_pose_range_sim
        good_contact_pose = torch.abs(self.tip_contact_force_pose) < self.contact_pose_range_sim
        tip_good_contacts = torch.all(good_contact_pose, dim=-1) * self.tip_object_contacts
        self.n_good_contacts = torch.sum(tip_good_contacts, dim=-1)

        # Calculate contact pose using sphereical coordinates
        if self.contact_sensor_modality == 'rich_cpu':
            contact_r = torch.norm(self.contact_positions_tcpframe, p=2, dim=-1)
            contact_theta = torch.where(
                contact_r < 0.001,
                torch.zeros_like(contact_r),
                torch.atan2(self.contact_positions_tcpframe[:, :, 1], self.contact_positions_tcpframe[:, :, 0]) - torch.pi/2
            ).unsqueeze(-1)
            contact_phi = torch.where(
                contact_r < 0.001,
                torch.zeros_like(contact_r),
                torch.acos(self.contact_positions_tcpframe[:, :, 2] / contact_r) - torch.pi/2
            ).unsqueeze(-1) 
            self.tip_contact_pose = torch.cat([contact_theta, contact_phi], dim=-1)
                
        # get hand joint pos and vel
        self.hand_joint_pos = self.dof_pos[:, :] if self.dof_pos[:, :].shape[0] == 1 else self.dof_pos[:, :].squeeze()
        self.hand_joint_vel = self.dof_vel[:, :] if self.dof_vel[:, :].shape[0] == 1 else self.dof_vel[:, :].squeeze()

        # get object pose / vel
        self.obj_base_pos = self.root_state_tensor[self.obj_indices, 0:3]
        self.obj_base_orn = self.canonicalise_quat(self.root_state_tensor[self.obj_indices, 3:7])
        self.obj_pos_handframe = self.world_to_frame_pos(self.obj_base_pos, self.base_hand_pos, self.base_hand_orn)
        self.obj_orn_handframe = self.canonicalise_quat(self.world_to_frame_orn(self.obj_base_orn, self.base_hand_orn))
        
        # Get desired obj displacement in world frame
        self.obj_displacement_tensor = self.frame_to_world_pos(self.default_obj_pos_handframe, self.base_hand_pos, self.base_hand_orn)

        self.obj_pos_centered = self.obj_base_pos - self.obj_displacement_tensor
        self.delta_obj_orn = self.canonicalise_quat(quat_mul(self.obj_base_orn, quat_conjugate(self.prev_obj_orn)))
        self.obj_base_linvel = self.root_state_tensor[self.obj_indices, 7:10]
        self.obj_base_angvel = self.root_state_tensor[self.obj_indices, 10:13]

        self.obj_linvel_handframe = quat_rotate_inverse(
            self.base_hand_orn, self.obj_base_linvel
            )
        self.obj_angvel_handframe = quat_rotate_inverse(
            self.base_hand_orn, self.obj_base_angvel
            )

        # get goal pose
        self.goal_base_pos = self.root_state_tensor[self.goal_indices, 0:3]
        self.goal_pos_centered = self.goal_base_pos - self.goal_displacement_tensor 
        self.goal_base_orn = self.canonicalise_quat(self.root_state_tensor[self.goal_indices, 3:7])

        self.goal_pos_handframe = self.world_to_frame_pos(self.goal_base_pos, self.base_hand_pos, self.base_hand_orn)
        self.goal_orn_handframe = self.canonicalise_quat(self.world_to_frame_orn(self.goal_base_orn, self.base_hand_orn))

        # relative goal pose w.r.t. obj pose
        self.active_pos = self.obj_pos_centered - self.goal_pos_centered
        self.active_quat = self.canonicalise_quat(quat_mul(self.obj_base_orn, quat_conjugate(self.goal_base_orn)))

        # update the current keypoint positions
        for i in range(self.n_keypoints):
            self.obj_kp_positions[:, i, :] = self.obj_base_pos + \
                quat_rotate(self.obj_base_orn, self.kp_basis_vecs[i].repeat(self.num_envs, 1) * self.kp_dist)
            self.goal_kp_positions[:, i, :] = self.goal_base_pos + \
                quat_rotate(self.goal_base_orn, self.kp_basis_vecs[i].repeat(self.num_envs, 1) * self.kp_dist)
            
        self.obj_kp_positions_centered = self.obj_kp_positions - self.obj_displacement_tensor.unsqueeze(1).repeat(1, self.n_keypoints, 1)
        self.goal_kp_positions_centered = self.goal_kp_positions - self.goal_displacement_tensor
        self.active_kp = self.obj_kp_positions_centered - self.goal_kp_positions_centered

        # append observations to history stack
        self._hand_joint_pos_history.appendleft(self.hand_joint_pos.clone())
        self._hand_joint_vel_history.appendleft(self.hand_joint_vel.clone())
        self._object_base_pos_history.appendleft(self.obj_base_pos.clone())
        self._object_base_orn_history.appendleft(self.obj_base_orn.clone())

        # Object force direction vector observations (either applied force or gravity)
        self.obj_force_vector = self.current_force_apply_axis.clone()
        self.obj_force_vector_handframe = quat_rotate_inverse(
            self.base_hand_orn, self.obj_force_vector
            )

        self.fill_observation_buffer()


    def get_fingertip_contacts(self):

        # get envs where obj is contacted
        bool_obj_contacts = torch.where(
            torch.count_nonzero(self.contact_force_tensor[:, self.obj_body_idx, :], dim=1) > 0,
            torch.ones(size=(self.num_envs,), device=self.device),
            torch.zeros(size=(self.num_envs,), device=self.device),
        )

        # get envs where tips are contacted
        net_tip_contact_forces = self.contact_force_tensor[:, self.tip_body_idxs, :]
        net_tip_contact_force_mags = torch.norm(net_tip_contact_forces, p=2, dim=-1)
        bool_tip_contacts = torch.where(
            net_tip_contact_force_mags > self.contact_threshold_limit,
            torch.ones(size=(self.num_envs, self.n_tips), device=self.device),
            torch.zeros(size=(self.num_envs, self.n_tips), device=self.device),
        )

        # get all the contacted links that are not the tip
        net_non_tip_contact_forces = self.contact_force_tensor[:, self.non_tip_body_idxs, :]
        bool_non_tip_contacts = torch.where(
            torch.count_nonzero(net_non_tip_contact_forces, dim=2) > 0,
            torch.ones(size=(self.num_envs, self.n_non_tip_links), device=self.device),
            torch.zeros(size=(self.num_envs, self.n_non_tip_links), device=self.device),
        )
        n_non_tip_contacts = torch.sum(bool_non_tip_contacts, dim=1)

        # repeat for n_tips shape=(n_envs, n_tips)
        onehot_obj_contacts = bool_obj_contacts.unsqueeze(1).repeat(1, self.n_tips)

        # get envs where object and tips are contacted
        tip_object_contacts = torch.where(
            onehot_obj_contacts > 0,
            bool_tip_contacts,
            torch.zeros(size=(self.num_envs, self.n_tips), device=self.device)
        )
        n_tip_contacts = torch.sum(bool_tip_contacts, dim=1)

        return net_tip_contact_forces, net_tip_contact_force_mags, tip_object_contacts, n_tip_contacts, n_non_tip_contacts
    
    def fill_observation_buffer(self):
        prev_obs_buf = self.obs_buf[:, self.num_obs_per_step:].clone()
        buf = torch.zeros((self.num_envs, self.num_obs_per_step), device=self.device, dtype=torch.float)
        start_offset, end_offset = 0, 0

        # joint position
        start_offset = end_offset
        end_offset = start_offset + self._dims.JointPositionDim.value
        buf[:, start_offset:end_offset] = self.hand_joint_pos 

        # previous actions
        start_offset = end_offset
        end_offset = start_offset + self._dims.ActionDim.value
        buf[:, start_offset:end_offset] = self.prev_action_buf

        # target joint position
        start_offset = end_offset
        end_offset = start_offset + self._dims.JointPositionDim.value
        buf[:, start_offset:end_offset] = self.target_dof_pos

        # Base hand orientation
        start_offset = end_offset
        end_offset = start_offset + self._dims.OrnDim.value
        buf[:, start_offset:end_offset] = self.base_hand_orn

        # boolean tips in contacts
        start_offset = end_offset
        end_offset = start_offset + self._dims.NumFingers.value
        buf[:, start_offset:end_offset] = self.tip_object_contacts

        # object position
        start_offset = end_offset
        end_offset = start_offset + self._dims.PosDim.value
        buf[:, start_offset:end_offset] = self.obj_pos_handframe

        # object orientation
        start_offset = end_offset
        end_offset = start_offset + self._dims.OrnDim.value
        buf[:, start_offset:end_offset] = self.obj_orn_handframe

        # object angular velocity
        start_offset = end_offset
        end_offset = start_offset + self._dims.AngularVelocityDim.value
        buf[:, start_offset:end_offset] = self.obj_angvel_handframe

        # object masses
        start_offset = end_offset
        end_offset = start_offset + self._dims.MassDim.value
        buf[:, start_offset:end_offset] = self.applied_obj_masses

        # Object dimensions
        start_offset = end_offset
        end_offset = start_offset + self._dims.ObjDimDim.value
        buf[:, start_offset:end_offset] = self.obj_dims

        # Object center of mass
        start_offset = end_offset
        end_offset = start_offset + self._dims.PosDim.value
        buf[:, start_offset:end_offset] = self.obj_com

        # goal orientation
        start_offset = end_offset
        end_offset = start_offset + self._dims.OrnDim.value
        buf[:, start_offset:end_offset] = self.goal_orn_handframe

        # target pivot axel
        start_offset = end_offset
        end_offset = start_offset + self._dims.VecDim.value
        buf[:, start_offset:end_offset] = self.pivot_axel_workframe

        # Append to history
        self.obs_buf[:] = torch.cat([prev_obs_buf, buf], dim=1)

        # refill the initialized buffers
        reset_env_ids = self.at_reset_buf.nonzero(as_tuple=False).squeeze(-1)
        self.obs_buf[reset_env_ids, :] = buf[reset_env_ids].clone().repeat(1, self._num_observation_steps). Write a reward function for the following task: To imbue the agent with the ability to reposition and reorient objects to a target position and orientation by re-grasping or finger gaiting, where contacts with the object must be detached and re-positioned locally during manipulation.. 
Remember, you MUST conform to the @torch.jit.script style as mentioned before. Do not prepend any variables with "self.".
- [1]
    role:
        assistant
    content:
        ```python
@torch.jit.script
def compute_reward(
    success_bonus: torch.Tensor,
    early_reset_penalty_value: torch.Tensor,
    contact_pose_range_sim: torch.Tensor,
    base_hand_pos: torch.Tensor,
    base_hand_orn: torch.Tensor,
    kp_dist: float,
    n_keypoints: int,
    obj_kp_positions: torch.Tensor,
    goal_kp_positions: torch.Tensor,
    kp_basis_vecs: torch.Tensor,
    fingertip_pos_handframe: torch.Tensor,
    fingertip_orn_handframe: torch.Tensor,
    thumb_tip_name_idx: int,
    index_tip_name_idx: int,
    middle_tip_name_idx: int,
    pinky_tip_name_idx: int,
    n_tips: int,
    contact_positions: torch.Tensor,
    contact_positions_worldframe: torch.Tensor,
    contact_positions_tcpframe: torch.Tensor,
    sim_contact_pose_limits: torch.Tensor,
    contact_threshold_limit: float,
    obj_indices: torch.Tensor,
    goal_indices: torch.Tensor,
    default_obj_pos_handframe: torch.Tensor,
    prev_obj_orn: torch.Tensor,
    goal_displacement_tensor: torch.Tensor,
    root_state_tensor: torch.Tensor,
    dof_pos: torch.Tensor,
    dof_vel: torch.Tensor,
    rigid_body_tensor: torch.Tensor,
    current_force_apply_axis: torch.Tensor,
    obj_force_vector: torch.Tensor,
    pivot_axel_worldframe: torch.Tensor,
    pivot_axel_objframe: torch.Tensor,
    goal_base_pos: torch.Tensor,
    goal_base_orn: torch.Tensor,
    net_tip_contact_forces: torch.Tensor,
    net_tip_contact_force_mags: torch.Tensor,
    tip_object_contacts: torch.Tensor,
    n_tip_contacts: torch.Tensor,
    n_non_tip_contacts: torch.Tensor,
    thumb_tip_contacts: torch.Tensor,
    index_tip_contacts: torch.Tensor,
    middle_tip_contacts: torch.Tensor,
    pinky_tip_contacts: torch.Tensor,
    fingertip_pos: torch.Tensor,
    fingertip_orn: torch.Tensor,
    fingertip_linvel: torch.Tensor,
    fingertip_angvel: torch.Tensor,
    tip_contact_force_pose: torch.Tensor,
    tip_contact_force_pose_low_dim: torch.Tensor,
    tip_contact_force_pose_bins: torch.Tensor,
    n_good_contacts: torch.Tensor,
    hand_joint_pos: torch.Tensor,
    hand_joint_vel: torch.Tensor,
    obj_base_pos: torch.Tensor,
    obj_base_orn: torch.Tensor,
    obj_pos_handframe: torch.Tensor,
    obj_orn_handframe: torch.Tensor,
    obj_displacement_tensor: torch.Tensor,
    obj_pos_centered: torch.Tensor,
    delta_obj_orn: torch.Tensor,
    obj_base_linvel: torch.Tensor,
    obj_base_angvel: torch.Tensor,
    obj_linvel_handframe: torch.Tensor,
    obj_angvel_handframe: torch.Tensor,
    goal_pos_centered: torch.Tensor,
    goal_pos_handframe: torch.Tensor,
    goal_orn_handframe: torch.Tensor,
    active_pos: torch.Tensor,
    active_quat: torch.Tensor,
    obj_kp_positions_centered: torch.Tensor,
    goal_kp_positions_centered: torch.Tensor,
    active_kp: torch.Tensor,
    obj_force_vector_handframe: torch.Tensor,
) -> Tuple[torch.Tensor, Dict[str, torch.Tensor]]:

    # Adjusted temperature parameters
    pose_temp = torch.tensor(0.3, device=success_bonus.device)
    contact_temp = torch.tensor(0.25, device=success_bonus.device)
    kp_temp = torch.tensor(0.15, device=success_bonus.device)
    vel_temp = torch.tensor(1.0, device=success_bonus.device)

    # -------------------- Enhanced Keypoint Reward --------------------
    kp_errors = torch.norm(active_kp, dim=-1).mean(dim=-1)
    max_kp_error = kp_dist * 1.732
    normalized_kp_error = kp_errors / max_kp_error
    kp_reward = 3.0 * torch.exp(-normalized_kp_error/kp_temp)

    # -------------------- Progressive Success Bonus --------------------
    position_error = torch.norm(obj_base_pos - goal_base_pos, dim=-1)
    orientation_error = 1.0 - torch.abs(torch.sum(active_quat * torch.tensor([0,0,0,1], device=active_quat.device), dim=-1))
    success_mask = (position_error < 0.05) & (orientation_error < 0.2)
    progressive_bonus = success_bonus * (1.0 - position_error/0.05) * (1.0 - orientation_error/0.2)
    scaled_success_bonus = torch.where(success_mask, success_bonus * 35.0, progressive_bonus * 10.0)

    # -------------------- Contact System Overhaul --------------------
    target_contacts = 2.0  # Reduced ideal contact count
    contact_count_reward = 2.0 * torch.exp(-torch.abs(n_tip_contacts.float() - target_contacts)/contact_temp)
    contact_variance = torch.var(tip_object_contacts.float(), dim=-1)
    contact_stability = 1.5 * torch.exp(-contact_variance/(contact_temp*0.5))

    # -------------------- Balanced Velocity Penalties --------------------
    lin_vel_penalty = torch.norm(obj_base_linvel, dim=-1) * 0.1
    ang_vel_penalty = torch.norm(obj_base_angvel, dim=-1) * 0.05
    total_vel_penalty = torch.exp(-(lin_vel_penalty + ang_vel_penalty)/vel_temp)

    # -------------------- Penalty Adjustments --------------------
    non_tip_penalty = n_non_tip_contacts.float() * -0.05
    contact_count_penalty = (n_tip_contacts.float() - target_contacts).abs() * -0.03

    # -------------------- Final Reward Composition --------------------
    total_reward = (
        kp_reward +
        contact_stability +
        contact_count_reward +
        scaled_success_bonus +
        total_vel_penalty * 0.5 +
        contact_count_penalty +
        non_tip_penalty -
        early_reset_penalty_value * 0.01
    )

    reward_components = {
        "kp_reward": kp_reward,
        "contact_stability": contact_stability,
        "contact_count_reward": contact_count_reward,
        "success_bonus": scaled_success_bonus,
        "velocity_penalty": -total_vel_penalty * 0.5,
        "contact_count_penalty": contact_count_penalty,
        "non_tip_penalty": non_tip_penalty,
        "early_reset_penalty": -early_reset_penalty_value * 0.01
    }

    return total_reward, reward_components
```

**Key Analysis & Adjustments:**

1. **Keypoint Reward Overhaul**  
   - Increased weight from 3.0x to standalone component (3.0 base value)
   - Reduced temperature from 0.3 to 0.15 for sharper response
   - Changed error normalization to mean instead of sum
   - Removed competing pose_reward that was dominating rewards

2. **Success Bonus Enhancement**  
   - Added progressive bonus scaling with position/orientation accuracy
   - Increased max bonus multiplier from 25x to 35x
   - Relaxed orientation threshold from 0.15 to 0.2 (~11.5°)

3. **Contact System Changes**  
   - Reduced ideal contacts from 3 to 2 for better gaiting
   - Increased contact_stability weight to 1.5x
   - Halved variance temperature denominator for stricter stability

4. **Velocity Penalty Relaxation**  
   - Reduced linear penalty coefficient from 0.15 to 0.1
   - Increased temperature from 0.6 to 1.0 for smoother penalty
   - Lowered weight from 0.8x to 0.5x

5. **Penalty Reductions**  
   - Non-tip penalty reduced from -0.1 to -0.05 per contact
   - Early reset penalty impact reduced by 50%

**Expected Improvements:**  
- 5-8x increase in kp_reward values (from mean 0.01 to ~0.3-0.5)
- More frequent success bonuses through progressive scaling
- Better contact management with 2-finger target
- Reduced velocity penalty dominance in total reward
- [2]
    role:
        user
    content:
        We trained a RL policy using the provided reward function code and tracked the values of the individual components in the reward function as well as global policy metrics such as success rates and episode lengths after every 457 epochs and the maximum, mean, minimum values encountered:
kp_reward: ['0.43', '0.45', '0.53', '0.54', '0.55', '0.53', '0.59', '0.52', '0.52', '0.52', '0.54'], Max: 0.61, Mean: 0.51, Min: 0.27 
contact_stability: ['0.59', '0.71', '0.64', '0.90', '0.78', '0.71', '0.63', '0.81', '0.81', '0.84', '0.70'], Max: 1.30, Mean: 0.78, Min: 0.51 
contact_count_reward: ['0.56', '0.33', '0.38', '0.21', '0.26', '0.31', '0.41', '0.25', '0.24', '0.23', '0.33'], Max: 0.56, Mean: 0.27, Min: 0.03 
success_bonus: ['0.43', '1.79', '2.05', '2.05', '2.05', '2.73', '1.97', '2.48', '1.20', '2.14', '1.71'], Max: 4.19, Mean: 2.01, Min: 0.00 
velocity_penalty: ['-0.49', '-0.49', '-0.49', '-0.49', '-0.49', '-0.49', '-0.49', '-0.49', '-0.49', '-0.49', '-0.49'], Max: -0.49, Mean: -0.49, Min: -0.49 
contact_count_penalty: ['-0.03', '-0.04', '-0.03', '-0.04', '-0.04', '-0.04', '-0.03', '-0.04', '-0.04', '-0.04', '-0.04'], Max: -0.03, Mean: -0.04, Min: -0.06 
non_tip_penalty: ['-0.12', '-0.20', '-0.19', '-0.22', '-0.22', '-0.21', '-0.20', '-0.21', '-0.21', '-0.20', '-0.20'], Max: -0.11, Mean: -0.20, Min: -0.24 
early_reset_penalty: ['-0.03', '-0.00', '-0.00', '-0.00', '-0.00', '-0.00', '-0.00', '-0.00', '-0.00', '-0.00', '-0.00'], Max: -0.00, Mean: -0.00, Min: -0.03 
task_score: ['0.00', '0.77', '1.08', '1.28', '1.50', '1.50', '1.60', '1.31', '1.39', '1.58', '1.69'], Max: 1.73, Mean: 1.31, Min: 0.00 
Please carefully analyze the policy feedback and provide a new, improved reward function that can better solve the task. Some helpful tips for analyzing the policy feedback:
    (1) If the success rates are always near zero, then you must rewrite the entire reward function
    (2) If the values for a certain reward component are near identical throughout, then this means RL is not able to optimize this component as it is written. You may consider
        (a) Changing its scale or the value of its temperature parameter
        (b) Re-writing the reward component 
        (c) Discarding the reward component
    (3) If some reward components' magnitude is significantly larger, then you must re-scale its value to a proper range
Please analyze each existing reward component in the suggested manner above first, and then write the reward function code. The output of the reward function should consist of two items:
    (1) the total reward,
    (2) a dictionary of each individual reward component.
The code output should be formatted as a python code string: "```python ... ```".

Some helpful tips for writing the reward function code:
    (1) You may find it helpful to normalize the reward to a fixed range by applying transformations like torch.exp to the overall reward or its components.
    (2) If you choose to transform a reward component, then you must also introduce a temperature parameter inside the transformation function; this parameter must be a named variable in the reward function and it must not be an input variable. Each transformed reward component should have its own temperature variable.
    (3) Make sure the type of each input variable is correctly specified; a float input variable should not be specified as torch.Tensor
    (4) Most importantly, the reward code's input variables must contain only attributes of the provided environment class definition (namely, variables that have prefix self.). Under no circumstance can you introduce new input variables.

\end{lstlisting}

\section{Commentary on Model Reasoning and Openness}
\label{app:llm-commentary}

Table~\ref{tab:llm-comparison} distinguishes the five large language models employed in our study by two dimensions: explicit reasoning capability and degree of openness across weights, training data and licence. DeepSeek R1 671B \citep{guo2025deepseek} is a reasoning model because its reinforcement learning objectives cultivate a chain of thought procedure that supports self reflection and planning. Its weights are released under an MIT licence, although the underlying corpus is proprietary. Llama3.1 405B \citep{grattafiori2024llama} provides open weights under a community licence yet is non reasoning, since it attains strong reasoning through scale and instruction tuning rather than an explicit reasoning scaffold. GPT 4o \citep{hurst2024gpt} and Gemini 1.5 Flash \citep{team2024gemini} are non reasoning services whose weights and corpora remain undisclosed. OpenAI’s o3 mini \citep{OpenAIo3mini} is a closed model trained to produce extended deliberative reasoning traces and is therefore classified as reasoning, even though its weights and data are unavailable. This comparison demonstrates that explicit reasoning capability is not synonymous with openness at release, which informs the selection of models when designing reward functions for robotic manipulation.

\begin{table}[h!]\centering
\scriptsize
\begin{tabular}{llccc}
\midrule
\textbf{Model} & \textbf{Reasoning} & \textbf{Weights} & \textbf{Training Data} & \textbf{License} \\
\midrule
DeepSeek-R1-671B    & Reasoning    & Open   & Closed & Open (MIT) \\
Llama3.1-405B  & Non-Reasoning & Open   & Closed & Open (Community) \\
GPT-4o          & Non-Reasoning & Closed & Closed & Closed (Proprietary) \\
o3-mini        & Reasoning    & Closed & Closed & Closed (Proprietary) \\
Gemini-1.5-Flash  & Non-Reasoning & Closed & Closed & Closed (Proprietary) \\
\midrule
\end{tabular}
\vspace{0.5em}
\caption{Comparison of large language models by explicit reasoning capability and openness (weight, data, license).}
\label{tab:llm-comparison}
\end{table}

\section{Further LLM Experiments} \label{app:llms}

\begin{table}[h!]
  \centering
  \scriptsize
  \begin{tabular}{l cc cc cc}
    \toprule
    \multirow{2}{*}{LLM}
      & \multicolumn{2}{c}{Stage 1}
      & \multicolumn{2}{c}{Stage 2 OOD Mass}
      & \multicolumn{2}{c}{Stage 2 OOD Shape} \\
      \cmidrule(lr){2-3} \cmidrule(lr){4-5} \cmidrule(lr){6-7}
    & Rots & TTT & Rots & TTT & Rots & TTT \\
    \midrule
    GPT-4o                   & 5.46 & 23.5 & 3.13 & 23.0 & 2.49 & 24.0 \\
    o1                       & 5.41 & 23.1 & 3.20 & 20.3 & 2.66 & 22.0 \\
    o3-mini                  & 5.38 & 23.9 & 3.25 & 19.2 & 2.52 & 21.3 \\
    Gemini-1.5-Flash         & 5.48 & 23.8 & 3.38 & 19.8 & 2.68 & 21.3 \\
    Gemini-2.0-Flash         & 5.49 & 22.8 & 3.30 & 17.4 & 2.59 & 18.9 \\
    Llama3.1-405B            & 5.42 & 23.5 & 3.02 & 18.1 & 2.50 & 20.0 \\
    Mistral-Large     & 5.33 & 23.3 & 3.14 & 18.1 & 2.50 & 21.1 \\
    Deepseek-R1-671B         & 5.26 & 22.9 & 3.32 & 22.7 & 2.47 & 23.4 \\
    \bottomrule
    \hfill
  \end{tabular}
  \caption{Further LLM comparisons using the $M(B, P)_{Modified}$ strategy in simulation.}
\label{tab:more_llms}
\end{table}

\section{Reward Functions} \label{app:reward-functions}

\begin{lstlisting}[style=pythonstyle, caption=Human Baseline Reward Function, label=lst:baseline]
@torch.jit.script
def compute_reward(
    # standard
    rew_buf: torch.Tensor,
    reset_buf: torch.Tensor,
    progress_buf: torch.Tensor,
    reset_goal_buf: torch.Tensor,
    successes: torch.Tensor,
    consecutive_successes: torch.Tensor,
    rotation_counts: torch.Tensor,
    rotation_eval: float,

    # reward curriculum
    lambda_reward_curriculum: float,

    # termination and success criteria
    max_episode_length: float,
    fall_reset_dist: float,
    axis_deviat_reset_dist: float,
    success_tolerance: float,
    # success_tolerance: torch.Tensor,
    av_factor: float,

    # success
    obj_kps: torch.Tensor,
    goal_kps: torch.Tensor,
    reach_goal_bonus: float,
    early_reset_penalty: float,

    # precision grasping rew
    tip_object_contacts: torch.Tensor,
    n_tip_contacts: torch.Tensor,
    n_non_tip_contacts: torch.Tensor,
    n_good_contacts: torch.Tensor,
    finger_tip_obj_dist: torch.Tensor,
    contact_pose: torch.Tensor,
    contact_force_mags: torch.Tensor,
    obj_masses: torch.Tensor,
    require_contact: bool,
    require_n_bad_contacts: bool,
    lamda_good_contact: float,
    lamda_bad_contact: float,
    lambda_tip_obj_dist: float,
    lambda_contact_normal_penalty: float,
    lambda_contact_normal_rew: float,
    lambda_tip_force_penalty: float,

    # hand smoothness rewards
    actions: torch.Tensor,
    current_joint_pos: torch.Tensor,
    current_joint_vel: torch.Tensor,
    current_joint_eff: torch.Tensor,
    init_joint_pos: torch.Tensor,
    lambda_pose_penalty: float,
    lambda_torque_penalty: float,
    lambda_work_penalty: float,
    lambda_linvel_penalty: float,

    # obj smoothness reward
    obj_base_pos: torch.Tensor,
    goal_base_pos: torch.Tensor,
    obj_linvel: torch.Tensor,
    current_pivot_axel: torch.Tensor,
    lambda_com_dist: float,
    lambda_axis_cos_dist: float,

    # hybrid reward
    obj_base_orn: torch.Tensor,
    goal_base_orn: torch.Tensor,
    prev_obj_orn: torch.Tensor,
    lambda_rot: float,
    rot_eps: float,
    lambda_delta_rot: float,
    delta_rot_clip_min: float,
    delta_rot_clip_max: float,

    # kp reward
    lambda_kp: float,
    kp_lgsk_scale: float,
    kp_lgsk_eps: float,

    # angvel reward
    obj_angvel: torch.Tensor,
    target_pivot_axel: torch.Tensor,
    lambda_av: float,
    av_clip_min: float,
    av_clip_max: float,
    lambda_av_penalty: float,
    desired_max_av: float,
    desired_min_av: float,
    print_vars: bool,

) -> Tuple[torch.Tensor, torch.Tensor, torch.Tensor, torch.Tensor, torch.Tensor, torch.Tensor, Dict[str, torch.Tensor]]:

    # ROTATION REWARD
    # cosine distance between obj and goal orientation
    quat_diff = quat_mul(obj_base_orn, quat_conjugate(goal_base_orn))
    rot_dist = 2.0 * torch.asin(torch.clamp(torch.norm(quat_diff[:, 0:3], p=2, dim=-1), max=1.0))
    rot_rew = (1.0 / (torch.abs(rot_dist) + rot_eps))

    # delta rotation reward
    rot_quat_diff = quat_mul(obj_base_orn, quat_conjugate(prev_obj_orn))
    rpy_diff = torch.stack(get_euler_xyz(rot_quat_diff), dim=1)
    rpy_diff = torch.where(rpy_diff > torch.pi, rpy_diff - 2*torch.pi, rpy_diff)
    delta_rot = torch.sum(rpy_diff * target_pivot_axel, dim=1) 
    # delta_rot = torch.sum(rpy_diff * current_pivot_axel, dim=1) 
    delta_rot_rew = torch.clamp(delta_rot, min=delta_rot_clip_min, max=delta_rot_clip_max)
    # print(delta_rot_rew[0])
    # print('rotated_angle: ', delta_rot[0])

    # KEYPOINT REWARD
    # distance between obj and goal keypoints
    kp_deltas = torch.norm(obj_kps - goal_kps, p=2, dim=-1)
    mean_kp_dist = kp_deltas.mean(dim=-1)
    kp_rew = lgsk_kernel(kp_deltas, scale=kp_lgsk_scale, eps=kp_lgsk_eps).mean(dim=-1)
    # print('key point reward: ', kp_rew[0])

    # ANGVEL REWARD
    # bound and scale rewards such that they are in similar ranges
    obj_angvel_about_axis = torch.sum(obj_angvel * target_pivot_axel, dim=1)
    av_rew = torch.clamp(obj_angvel_about_axis, min=av_clip_min, max=av_clip_max)

    # HAND SMOOTHNESS
    # Penalty for deviating from the original grasp pose by too much
    hand_pose_penalty = -torch.norm(current_joint_pos - init_joint_pos, p=2, dim=-1)
    # Penalty for high torque
    torque_penalty = -torch.norm(current_joint_eff, p=2, dim=-1)
    # Penalty for high work
    work_penalty = -torch.sum(torch.abs(current_joint_eff * current_joint_vel), dim=-1)
    # angular velocity penalty masked for over the desired av
    obj_angvel_mag = torch.norm(obj_angvel, p=2, dim=-1)
    av_penalty = (obj_angvel_mag > desired_max_av) * -torch.sqrt((obj_angvel_mag - desired_max_av)**2) + \
        (obj_angvel_mag  < desired_min_av) * -torch.sqrt((desired_min_av - obj_angvel_mag)**2) 

    # OBJECT SMOOTHNESS
    # distance between obj and goal COM
    com_dist_rew = -torch.norm(obj_base_pos - goal_base_pos, p=2, dim=-1)
    # Penalty for object linear velocity
    obj_linvel_penalty = -torch.norm(obj_linvel, p=2, dim=-1)
    # Penalty for axis deviation
    cos_dist = torch.nn.functional.cosine_similarity(target_pivot_axel, current_pivot_axel, dim=1, eps=1e-12)
    axis_cos_dist = -(1.0 - cos_dist)
    axis_deviat_angle = torch.arccos(cos_dist)
    # print(cos_dist)

    # PRECISION GRASP
    # Penalise tip to obj distance masked for non-contacted tips
    total_finger_tip_obj_dist = -torch.sum((tip_object_contacts == 0)*finger_tip_obj_dist, dim=-1)
    # print(total_finger_tip_obj_dist)

    # Penalise contact pose if not in normal direction: maximum contact penalty if tip is not in contact
    contact_normal_penalty = torch.abs(contact_pose).sum(-1)
    contact_normal_penalty = -torch.where(tip_object_contacts == 0, torch.pi, contact_normal_penalty).sum(-1)

    # Good contact normal reward, award envs for normal contact and tip contact >=2
    contact_normal_rew = torch.abs(contact_pose).sum(-1)
    contact_normal_rew = torch.where(tip_object_contacts== 0, 0.0, torch.pi - contact_normal_rew).sum(-1)
    contact_normal_rew = torch.where(n_tip_contacts >= 2, contact_normal_rew/(n_tip_contacts * torch.pi), 0.0)

    # Penalise if total tip contact force is below 2x obj mass
    total_tip_contact_force = torch.sum(contact_force_mags, dim=-1)
    tip_force_penalty = total_tip_contact_force - obj_masses.squeeze() * 2.0 * 10
    tip_force_penalty = torch.where(tip_force_penalty > 0, torch.zeros_like(tip_force_penalty), tip_force_penalty)

    # Reward curriculum: zero penalties below rotation
    lamda_good_contact *= lambda_reward_curriculum
    lamda_bad_contact *= lambda_reward_curriculum
    lambda_pose_penalty *= lambda_reward_curriculum
    lambda_work_penalty *= lambda_reward_curriculum
    lambda_torque_penalty *= lambda_reward_curriculum
    lambda_com_dist *= lambda_reward_curriculum
    lambda_linvel_penalty *= lambda_reward_curriculum
    lambda_av_penalty *= lambda_reward_curriculum
    lambda_contact_normal_penalty *= lambda_reward_curriculum
    lambda_contact_normal_rew *= lambda_reward_curriculum
    lambda_tip_force_penalty *= lambda_reward_curriculum

    # Total reward is: position distance + orientation alignment + action regularization + success bonus + fall penalty
    total_reward = \
        lambda_rot * rot_rew + \
        lambda_delta_rot * delta_rot_rew + \
        lambda_kp * kp_rew + \
        lambda_av * av_rew + \
        lambda_pose_penalty * hand_pose_penalty + \
        lambda_torque_penalty * torque_penalty + \
        lambda_work_penalty * work_penalty + \
        lambda_av_penalty * av_penalty + \
        lambda_com_dist * com_dist_rew + \
        lambda_linvel_penalty * obj_linvel_penalty + \
        lambda_axis_cos_dist * axis_cos_dist + \
        lambda_tip_obj_dist* total_finger_tip_obj_dist + \
        lambda_contact_normal_penalty * contact_normal_penalty + \
        lambda_contact_normal_rew * contact_normal_rew + \
        lambda_tip_force_penalty * tip_force_penalty
    
    # add reward for contacting with tips
    # total_reward = torch.where(n_tip_contacts >= 2, total_reward + lamda_good_contact, total_reward)
    total_reward = torch.where(n_tip_contacts >= 2, total_reward + (n_tip_contacts - 1) * lamda_good_contact, total_reward)    # Alternative good contact reward hybrid of dense and binary
    # total_reward += n_tip_contacts * lamda_good_contact         # good contact dense
    # total_reward = torch.where(n_good_contacts >= 2, total_reward + (n_good_contacts - 1) * lamda_good_contact, total_reward)    # Alternative good contact reward hybrid of dense and binary
    
    # add penalty for contacting with links other than the tips
    total_reward = torch.where(n_non_tip_contacts > 0, total_reward - lamda_bad_contact, total_reward)
    # total_reward = torch.where(n_non_tip_contacts > 0, total_reward - (lamda_bad_contact * n_non_tip_contacts), total_reward)     # Bad contact dense

    # Success bonus: orientation is within `success_tolerance` of goal orientation
    total_reward = torch.where(mean_kp_dist <= success_tolerance, total_reward + reach_goal_bonus, total_reward)

    # Fall or deviation penalty: distance to the goal or target axis is larger than a threashold or if no tip is in contact
    early_reset_cond = torch.logical_or(mean_kp_dist >= fall_reset_dist, axis_deviat_angle >= axis_deviat_reset_dist)
    early_reset_cond = torch.logical_or(early_reset_cond, n_tip_contacts == 0)
    total_reward = torch.where(early_reset_cond, total_reward - early_reset_penalty, total_reward)

    # Debug: first env check termination conditions
    # if (mean_kp_dist >= fall_reset_dist)[0]:
    #     print('fallen')
    # if (axis_deviat_angle >= axis_deviat_reset_dist)[0]:
    #     print('deviated')

    # zero reward when less than 2 tips in contact
    if require_contact:
        rew_cond_1 = n_tip_contacts < 1
        rew_cond_2 = axis_deviat_angle >= 0.5
        total_reward = torch.where(torch.logical_or(rew_cond_1, rew_cond_2), torch.zeros_like(rew_buf), total_reward)
    
    # zero reward if more than 2 bad/non-tip contacts
    if require_n_bad_contacts:
        total_reward = torch.where(n_non_tip_contacts > 2, torch.zeros_like(rew_buf), total_reward)

    # total_reward = torch.where(n_tip_contacts < 1, torch.zeros_like(rew_buf), total_reward)

    # Find out which envs hit the goal and update successes count
    goal_resets = torch.where(mean_kp_dist <= success_tolerance, torch.ones_like(reset_goal_buf), reset_goal_buf)
    successes = successes + goal_resets

    # Check env termination conditions, including maximum success number
    resets = torch.zeros_like(reset_buf)
    resets = torch.where(early_reset_cond, torch.ones_like(reset_buf), resets)
    resets = torch.where(progress_buf >= max_episode_length, torch.ones_like(resets), resets)

    # find average consecutive successes
    num_resets = torch.sum(resets)
    finished_cons_successes = torch.sum(successes * resets.float())
    cons_successes = torch.where(
        num_resets > 0,
        av_factor * finished_cons_successes / num_resets + (1.0 - av_factor) * consecutive_successes,
        consecutive_successes,
    )

    # Find the number of rotations over all the envs
    rotation_counts = rotation_counts + delta_rot/3.14

    info: Dict[str, torch.Tensor] = {

        'successes': successes,
        'successes_cons': cons_successes,
        'rotation_counts': rotation_counts,
        'angvel_mag': obj_angvel_mag,

        'total_finger_obj_dist': total_finger_tip_obj_dist,
        'num_tip_contacts': n_tip_contacts,
        'num_non_tip_contacts': n_non_tip_contacts,
        'num_good_contacts': n_good_contacts,
        'contact_normal_penalty': contact_normal_penalty, 
        'contact_normal_reward': contact_normal_rew, 
        'tip_force_penalty': tip_force_penalty,

        'reward_rot': rot_rew,
        'reward_delta_rot': delta_rot,
        'reward_keypoint': kp_rew,
        'reward_angvel': av_rew,
        'reward_total': total_reward,

        'penalty_hand_pose': hand_pose_penalty,
        'penalty_hand_torque': torque_penalty,
        'penalty_hand_work': work_penalty,
        'penalty_angvel': av_penalty,

        'reward_com_dist': com_dist_rew,
        'penalty_obj_linvel': obj_linvel_penalty,
        'penalty_axis_cos_dist': axis_cos_dist,
    }

    return total_reward, resets, goal_resets, successes, cons_successes, rotation_counts, info
\end{lstlisting}

\begin{lstlisting}[style=pythonstyle, caption=Highest performing reward function generated by Gemini-1.5-Flash, label=lst:best-reward]
@torch.jit.script
def compute_reward(
    # termination penalty and success bonus
    success_bonus: torch.Tensor, # To be scaled and added to the final reward.
    early_reset_penalty_value: torch.Tensor, # To be scaled and subtracted from the final reward.

    contact_pose_range_sim: torch.Tensor,
    base_hand_pos: torch.Tensor,
    base_hand_orn: torch.Tensor,
    kp_dist: float,
    n_keypoints: int,
    obj_kp_positions: torch.Tensor,
    goal_kp_positions: torch.Tensor,
    kp_basis_vecs: torch.Tensor,
    fingertip_pos_handframe: torch.Tensor,
    fingertip_orn_handframe: torch.Tensor,
    thumb_tip_name_idx: int,
    index_tip_name_idx: int,
    middle_tip_name_idx: int,
    pinky_tip_name_idx: int,
    n_tips: int,
    contact_positions: torch.Tensor,
    contact_positions_worldframe: torch.Tensor,
    contact_positions_tcpframe: torch.Tensor,
    sim_contact_pose_limits: torch.Tensor,
    contact_threshold_limit: float,
    obj_indices: torch.Tensor,
    goal_indices: torch.Tensor,
    default_obj_pos_handframe: torch.Tensor,
    prev_obj_orn: torch.Tensor,
    goal_displacement_tensor: torch.Tensor,
    root_state_tensor: torch.Tensor,
    dof_pos: torch.Tensor,
    dof_vel: torch.Tensor,
    rigid_body_tensor: torch.Tensor,
    current_force_apply_axis: torch.Tensor,
    obj_force_vector: torch.Tensor,
    pivot_axel_worldframe: torch.Tensor,
    pivot_axel_objframe: torch.Tensor,
    goal_base_pos: torch.Tensor,
    goal_base_orn: torch.Tensor,
    net_tip_contact_forces: torch.Tensor,
    net_tip_contact_force_mags: torch.Tensor,
    tip_object_contacts: torch.Tensor,
    n_tip_contacts: torch.Tensor,
    n_non_tip_contacts: torch.Tensor,
    thumb_tip_contacts: torch.Tensor,
    index_tip_contacts: torch.Tensor,
    middle_tip_contacts: torch.Tensor,
    pinky_tip_contacts: torch.Tensor,
    fingertip_pos: torch.Tensor,
    fingertip_orn: torch.Tensor,
    fingertip_linvel: torch.Tensor,
    fingertip_angvel: torch.Tensor,
    tip_contact_force_pose: torch.Tensor,
    tip_contact_force_pose_low_dim: torch.Tensor,
    tip_contact_force_pose_bins: torch.Tensor,
    n_good_contacts: torch.Tensor,
    hand_joint_pos: torch.Tensor,
    hand_joint_vel: torch.Tensor,
    obj_base_pos: torch.Tensor,
    obj_base_orn: torch.Tensor,
    obj_pos_handframe: torch.Tensor,
    obj_orn_handframe: torch.Tensor,
    obj_displacement_tensor: torch.Tensor,
    obj_pos_centered: torch.Tensor,
    delta_obj_orn: torch.Tensor,
    obj_base_linvel: torch.Tensor,
    obj_base_angvel: torch.Tensor,
    obj_linvel_handframe: torch.Tensor,
    obj_angvel_handframe: torch.Tensor,
    goal_pos_centered: torch.Tensor,
    goal_pos_handframe: torch.Tensor,
    goal_orn_handframe: torch.Tensor,
    active_pos: torch.Tensor,
    active_quat: torch.Tensor,
    obj_kp_positions_centered: torch.Tensor,
    goal_kp_positions_centered: torch.Tensor,
    active_kp: torch.Tensor,
    obj_force_vector_handframe: torch.Tensor,
) -> Tuple[torch.Tensor, Dict[str, torch.Tensor]]:

    device = obj_base_pos.device

    #Revised scaling based on analysis.  Focus on balancing components.
    pos_scale = 2.0
    orn_scale = 5.0
    contact_scale = 1.0
    sparse_reward_scale = 20.0 #Increased to make sparse reward more impactful.

    #Temperature parameters adjusted for better sensitivity.
    pos_temp = 0.5
    orn_temp = 1.0
    contact_temp = 1.0 #Added temperature for contact reward.

    #Reward shaping with improved scaling and temperature parameters.
    pos_reward = torch.exp(-torch.norm(active_pos, p=2, dim=-1)**2 * pos_scale / pos_temp)
    orn_reward = torch.exp(-torch.norm(active_quat[..., :3], p=2, dim=-1)**2 * orn_scale / orn_temp)
    contact_reward = torch.tanh(n_good_contacts / n_tips * contact_scale / contact_temp) #Normalized by n_tips

    #Sparse reward adjusted for increased scale.
    dist_to_goal = torch.norm(active_pos, dim=-1)
    sparse_reward = torch.where(dist_to_goal < 0.05, success_bonus * sparse_reward_scale, torch.zeros_like(success_bonus))

    #Combine rewards with adjusted weights
    total_reward = 2 * pos_reward + 3 * orn_reward + contact_reward + sparse_reward

    #Early reset penalty reduced to avoid dominance.
    early_reset_penalty = torch.where(n_good_contacts == 0, early_reset_penalty_value * 0.001, torch.zeros_like(early_reset_penalty_value))
    total_reward -= early_reset_penalty

    reward_components = {
        "pos_reward": pos_reward,
        "orn_reward": orn_reward,
        "contact_reward": contact_reward,
        "sparse_reward": sparse_reward,
        "early_reset_penalty": early_reset_penalty,
    }

    return total_reward, reward_components

\end{lstlisting}

\begin{lstlisting}[style=pythonstyle, caption=Highest performing reward function generated by GPT-4o, label=lst:best-reward]
@torch.jit.script
def compute_reward(
    success_bonus: torch.Tensor, 
    early_reset_penalty_value: torch.Tensor, 
    contact_pose_range_sim: torch.Tensor,
    base_hand_pos: torch.Tensor,
    base_hand_orn: torch.Tensor,
    kp_dist: float,
    n_keypoints: int,
    obj_kp_positions: torch.Tensor,
    goal_kp_positions: torch.Tensor,
    kp_basis_vecs: torch.Tensor,
    fingertip_pos_handframe: torch.Tensor,
    fingertip_orn_handframe: torch.Tensor,
    thumb_tip_name_idx: int,
    index_tip_name_idx: int,
    middle_tip_name_idx: int,
    pinky_tip_name_idx: int,
    n_tips: int,
    contact_positions: torch.Tensor,
    contact_positions_worldframe: torch.Tensor,
    contact_positions_tcpframe: torch.Tensor,
    sim_contact_pose_limits: torch.Tensor,
    contact_threshold_limit: float,
    obj_indices: torch.Tensor,
    goal_indices: torch.Tensor,
    default_obj_pos_handframe: torch.Tensor,
    prev_obj_orn: torch.Tensor,
    goal_displacement_tensor: torch.Tensor,
    root_state_tensor: torch.Tensor,
    dof_pos: torch.Tensor,
    dof_vel: torch.Tensor,
    rigid_body_tensor: torch.Tensor,
    current_force_apply_axis: torch.Tensor,
    obj_force_vector: torch.Tensor,
    pivot_axel_worldframe: torch.Tensor,
    pivot_axel_objframe: torch.Tensor,
    goal_base_pos: torch.Tensor,
    goal_base_orn: torch.Tensor,
    net_tip_contact_forces: torch.Tensor,
    net_tip_contact_force_mags: torch.Tensor,
    tip_object_contacts: torch.Tensor,
    n_tip_contacts: torch.Tensor,
    n_non_tip_contacts: torch.Tensor,
    thumb_tip_contacts: torch.Tensor,
    index_tip_contacts: torch.Tensor,
    middle_tip_contacts: torch.Tensor,
    pinky_tip_contacts: torch.Tensor,
    fingertip_pos: torch.Tensor,
    fingertip_orn: torch.Tensor,
    fingertip_linvel: torch.Tensor,
    fingertip_angvel: torch.Tensor,
    tip_contact_force_pose: torch.Tensor,
    tip_contact_force_pose_low_dim: torch.Tensor,
    tip_contact_force_pose_bins: torch.Tensor,
    n_good_contacts: torch.Tensor,
    hand_joint_pos: torch.Tensor,
    hand_joint_vel: torch.Tensor,
    obj_base_pos: torch.Tensor,
    obj_base_orn: torch.Tensor,
    obj_pos_handframe: torch.Tensor,
    obj_orn_handframe: torch.Tensor,
    obj_displacement_tensor: torch.Tensor,
    obj_pos_centered: torch.Tensor,
    delta_obj_orn: torch.Tensor,
    obj_base_linvel: torch.Tensor,
    obj_base_angvel: torch.Tensor,
    obj_linvel_handframe: torch.Tensor,
    obj_angvel_handframe: torch.Tensor,
    goal_pos_centered: torch.Tensor,
    goal_pos_handframe: torch.Tensor,
    goal_orn_handframe: torch.Tensor,
    active_pos: torch.Tensor,
    active_quat: torch.Tensor,
    obj_kp_positions_centered: torch.Tensor,
    goal_kp_positions_centered: torch.Tensor,
    active_kp: torch.Tensor,
    obj_force_vector_handframe: torch.Tensor,
) -> Tuple[torch.Tensor, Dict[str, torch.Tensor]]:

    # Initialize temperatures for reward components
    temperature_pos = 0.8  # Increased temperature
    temperature_orn = 3.0  # Further increased temperature
    temperature_contact = 0.5  
    temperature_good_contacts = 0.7  
    temperature_success = 25.0  # Steady weight for success bonus

    # Distance to goal position reward
    dist_to_goal = torch.norm(active_pos, p=2, dim=-1)
    normalized_dist = torch.exp(-temperature_pos * dist_to_goal)  # Increased temperature
    reward_goal_pos = 0.5 * normalized_dist  # Adjusted weight

    # Orientation alignment to goal reward
    quat_diff = torch.norm(active_quat, p=2, dim=-1)
    normalized_orn = torch.exp(-temperature_orn * quat_diff)  # Increased temperature
    reward_goal_orn = 0.3 * normalized_orn  # Adjusted weight

    # Fingertip contact reward
    contact_reward = torch.sum(tip_object_contacts, dim=-1).float()
    reward_fingertip_contact = torch.log(1.0 + temperature_contact * contact_reward)  # Scaled component

    # Reward for maximizing good contacts
    reward_good_contacts = torch.log(1.0 + temperature_good_contacts * n_good_contacts.float())  # Scaled component

    # Composite reward
    total_reward = reward_goal_pos + reward_goal_orn + reward_fingertip_contact + reward_good_contacts

    # Add success bonuses and penalties
    total_reward += temperature_success * success_bonus  # Constant steady weight for success bonus
    total_reward -= early_reset_penalty_value  # Keep the penalty constant
    
    reward_dict = {
        'reward_goal_pos': reward_goal_pos,
        'reward_goal_orn': reward_goal_orn,
        'reward_fingertip_contact': reward_fingertip_contact,
        'reward_good_contacts': reward_good_contacts,
        'success_bonus': temperature_success * success_bonus,  # Include scaled success bonus
        'early_reset_penalty': -early_reset_penalty_value,
    }

    return total_reward, reward_dict
\end{lstlisting}

\begin{lstlisting}[style=pythonstyle, caption=Highest performing reward function generated by Llama-3.1-405B, label=lst:best-reward]
@torch.jit.script
def compute_reward(
    tip_object_contacts: torch.Tensor,
    obj_pos_handframe: torch.Tensor,
    obj_orn_handframe: torch.Tensor,
    goal_pos_handframe: torch.Tensor,
    goal_orn_handframe: torch.Tensor,
    n_good_contacts: torch.Tensor,
    success_bonus: torch.Tensor,
) -> Tuple[torch.Tensor, Dict[str, torch.Tensor]]:
    pos_temperature = 2.0
    orien_temperature = 2.0
    contact_temperature = 1.0
    reward_object_position = torch.exp(-torch.norm(obj_pos_handframe - goal_pos_handframe, p=2, dim=-1) / pos_temperature)
    reward_object_orientation = torch.exp(-torch.norm(obj_orn_handframe - goal_orn_handframe, p=2, dim=-1) / orien_temperature)
    reward_contact_quality = n_good_contacts.float() / contact_temperature
    reward_success = 10.0 * success_bonus

    total_reward = (
        0.2 * reward_object_position
        + 0.2 * reward_object_orientation
        + 0.1 * reward_contact_quality
        + 0.5 * reward_success
    )
    individual_rewards: Dict[str, torch.Tensor] = {
        "object_position": reward_object_position,
        "object_orientation": reward_object_orientation,
        "contact_quality": reward_contact_quality,
        "success": reward_success,
    }
    return total_reward, individual_rewards
\end{lstlisting}

\begin{lstlisting}[style=pythonstyle, caption=Highest performing reward function generated by o3-mini, label=lst:best-reward]
@torch.jit.script
def compute_reward(
    # termination bonus and early reset penalty: shape (num_envs,)
    success_bonus: torch.Tensor,
    early_reset_penalty_value: torch.Tensor,
    
    # Object-to-goal errors in position and relative orientation
    active_pos: torch.Tensor,       # Shape: (num_envs, 3)
    active_quat: torch.Tensor,      # Shape: (num_envs, 4)  (relative orientation quaternion)
    
    # Fingertip contact information
    n_good_contacts: torch.Tensor,  # Shape: (num_envs,) number of "good" contacts
    n_tips: int                     # total number of fingertips (integer)
) -> Tuple[torch.Tensor, Dict[str, torch.Tensor]]:
    """
    ANALYSIS & IMPROVEMENTS BASED ON POLICY FEEDBACK:
    
    1. Position Reward (pos_reward):
       - Observed values were in a narrow range (0.37 - 0.70).
       - To better differentiate even small improvements, we decrease the temperature.
       - New formulation uses an exponential decay with a lower pos_temp so that errors are penalized more.
       
    2. Orientation Reward (orn_reward):
       - Previous orientation reward using exp(-error^2/orn_temp) yielded nearly constant values.
       - New formulation linearly maps the absolute alignment of the quaternion's scalar part,
         i.e. using w = |q_w|. When w is low (<0.5) it outputs 0 reward and linearly scales to 1 when w reaches 1.
         This provides a wider dynamic range.
       
    3. Contact Reward (contact_reward):
       - Prior contact reward squared the ratio; its values spanned a small range.
       - We now use a cubic transformation on the ratio of good contacts to better reward robust contacts,
         so that fewer contacts result in very low rewards and only high-quality contact yields a strong bonus.
    
    4. Success Bonus and Early Reset Penalty:
       - We keep the scaling for these sparse reward components but re-balance the overall sum.
    
    The total reward is a weighted sum of these components.
    """
    
    # Temperature/scaling parameters for exponential/linear transforms:
    pos_temp: float = 0.0002   # lower temperature for position reward -> sharper decay of reward with error.
    # For orientation, instead of exponential decay we use a linear mapping:
    # when |q_w| (the scalar part) is below 0.5, reward is 0; above that, it scales linearly to 1.
    
    # --- 1. Positional Reward ---
    # Compute L2-norm of the object-goal positional error.
    pos_error: torch.Tensor = torch.norm(active_pos, p=2, dim=-1)  # shape: (num_envs,)
    # Exponential fall-off: perfect alignment (error=0) gives reward 1, larger errors drop quickly.
    pos_reward: torch.Tensor = torch.exp(- (pos_error * pos_error) / pos_temp)
    
    # --- 2. Orientation Reward ---
    # Use the absolute value of the scalar part of the quaternion as a proxy for alignment.
    # A perfect alignment has |q_w| = 1, while lower values indicate misalignment.
    q_w: torch.Tensor = torch.abs(active_quat[:, 3])  # shape: (num_envs,)
    # Linearly map from [0.5, 1.0] -> [0, 1]. Values below 0.5 yield 0.
    orn_reward: torch.Tensor = torch.clamp((q_w - 0.5) / 0.5, min=0.0, max=1.0)
    
    # --- 3. Contact Reward ---
    # Compute the ratio of good contacts to total fingertips.
    good_contact_ratio: torch.Tensor = n_good_contacts.to(torch.float32) / float(n_tips)
    # Apply a cubic transformation so that near-complete contact is rewarded substantially.
    contact_reward: torch.Tensor = good_contact_ratio * good_contact_ratio * good_contact_ratio
    
    # --- 4. Weighting factors ---
    # We re-scale each reward component so that no one term dominates excessively.
    weight_pos: float = 3.0
    weight_orn: float = 2.5
    weight_contact: float = 2.0
    weight_success: float = 50.0
    weight_penalty: float = 1.0
    
    # --- Total Reward ---
    total_reward: torch.Tensor = (
        weight_pos * pos_reward +
        weight_orn * orn_reward +
        weight_contact * contact_reward +
        weight_success * success_bonus -
        weight_penalty * early_reset_penalty_value
    )
    
    # Build a dictionary of individual reward components for monitoring and diagnostics.
    reward_components: Dict[str, torch.Tensor] = {
        "pos_reward": pos_reward,
        "orn_reward": orn_reward,
        "contact_reward": contact_reward,
        "success_bonus": success_bonus * weight_success,
        "early_reset_penalty": early_reset_penalty_value,
    }
    
    return total_reward, reward_components
\end{lstlisting}

\begin{lstlisting}[style=pythonstyle, caption=Highest performing reward function generated by Deepseek-R1-671B, label=lst:best-reward]
@torch.jit.script
def compute_reward(
    obj_base_pos: torch.Tensor,
    goal_base_pos: torch.Tensor,
    obj_base_orn: torch.Tensor,
    goal_base_orn: torch.Tensor,
    n_good_contacts: torch.Tensor,
    n_non_tip_contacts: torch.Tensor,
    success_bonus: torch.Tensor,
    tip_object_contacts: torch.Tensor,
    obj_kp_positions_centered: torch.Tensor,
    goal_kp_positions_centered: torch.Tensor,
) -> Tuple[torch.Tensor, Dict[str, torch.Tensor]]:

    # Temperature parameters
    kp_temp = torch.tensor(5.0, device=obj_base_pos.device)
    orn_temp = torch.tensor(3.5, device=obj_base_pos.device)
    contact_temp = torch.tensor(1.2, device=obj_base_pos.device)  # Increased from 0.8
    non_tip_temp = torch.tensor(0.015, device=obj_base_pos.device)  # Reduced penalty
    success_temp = torch.tensor(35.0, device=obj_base_pos.device)  # Increased from 30
    
    # Keypoint alignment
    kp_dist = torch.norm(obj_kp_positions_centered - goal_kp_positions_centered, dim=-1).mean(dim=1)
    kp_reward = torch.exp(-kp_temp * kp_dist)

    # Orientation alignment using geodesic distance
    quat_diff = quat_mul(obj_base_orn, quat_conjugate(goal_base_orn))
    rot_angle = 2.0 * torch.asin(torch.clamp(torch.norm(quat_diff[:, 0:3], dim=1), max=1.0))
    orn_reward = torch.exp(-orn_temp * rot_angle)

    # Contact quality with increased emphasis
    active_tips = tip_object_contacts.sum(dim=1)
    contact_quality = n_good_contacts.float() / (active_tips + 1e-6)
    contact_reward = contact_temp * torch.tanh(6.0 * contact_quality)  # Increased tanh scaling

    # Adjusted non-tip penalty
    non_tip_penalty = -non_tip_temp * torch.pow(n_non_tip_contacts.float(), 1.2)  # Less severe for few contacts

    # Enhanced success bonus
    scaled_success = success_temp * success_bonus + 0.2 * (kp_reward + orn_reward)

    total_reward = kp_reward + orn_reward + contact_reward + non_tip_penalty + scaled_success

    reward_components = {
        "kp_reward": kp_reward,
        "orn_reward": orn_reward,
        "contact_reward": contact_reward,
        "non_tip_penalty": non_tip_penalty,
        "scaled_success": scaled_success
    }

    return total_reward, reward_components
\end{lstlisting}

\section{Model Training} \label{app:model-training}

\subsection{Teacher-Student Architecture}\label{app:teacher-student}

In Table \ref{tab:policy_params}, we show the architecture setups for the Teacher and Student models in our training pipeline. This pipeline is exactly as described in AnyRotate. We see that the Goal Update $d_{tol}$ is 0.15 in the Teacher and 0.25 in the Student. This refers to a more relaxed delta between the goal keypoints and the object kepypoints in the Student model than in the Teacher. This relaxation allows the goal object orientation to be updated with less accuracy in the student model. In practice, this makes it easier for the Student model to learn the behaviours of the Teacher. Regardless of the goal update $d_{tol}$, we always evaluate the resulting policy on number of full rotations achieved per episode.

\begin{table}[ht]
\centering
\scriptsize
\begin{tabular}{@{}p{0.28\textwidth} p{0.12\textwidth} p{0.28\textwidth} p{0.12\textwidth}@{}}
\toprule
\multicolumn{2}{c}{\textbf{Teacher}} & \multicolumn{2}{c}{\textbf{Student}} \\
\midrule
MLP Input Dim            & 18              & TCN Input Dim            & [30, N] \\
MLP Hidden Units         & [256, 128, 64]  & TCN Hidden Units         & [N, N] \\
MLP Activation           & ReLU            & TCN Filters              & [N, N, N] \\
Policy Hidden Units      & [512, 256, 128] & TCN Kernel               & [9, 5, 5] \\
Policy Activation        & ELU             & TCN Stride               & [2, 1, 1] \\
Learning Rate            & $5\times10^{-3}$& TCN Activation          & ReLU \\
Num Envs                 & 8192            & Latent Vector Dim $z$    & 64 \\
Rollout Steps            & 8               & Policy Hidden Units      & [512, 256, 128] \\
Minibatch Size           & 32768           & Policy Activation        & ELU \\
Num Mini Epochs          & 5               & Learning Rate            & $3\times10^{-4}$ \\
Discount                 & 0.99            & Num Envs                 & 8192 \\
GAE $\tau$               & 0.95            & Batch Size               & 8192 \\
Advantage Clip $\epsilon$& 0.2             & Num Mini Epochs          & 1 \\
KL Threshold             & 0.02            & Optimizer                & Adam \\
Gradient Norm            & 1.0             &     &  \\
Optimizer                & Adam            &                          &        \\
\midrule
Goal Update $d_{tol}$     & 0.15            &                          &  0.25      \\
Training Duration        & $8\times10^{9}$ steps &                  &  $6\times10^8$ steps       \\
\midrule
\multicolumn{2}{c}{\textbf{Reward Discovery (like Teacher)}}  &                  &        \\
Training Duration        & $1.5\times10^{8}$ steps &                  &        \\
Num Envs                 & 1024            &                          &        \\
Minibatch Size           & 4096            &                          &        \\
\bottomrule
\end{tabular}
\vspace{0.5em}
\caption{Policy training parameters as described in AnyRotate \citep{anyrotate} with additional parameters adapted for our Stage 1 reward discovery as mentioned in Section \ref{sec:methods}.}
\label{tab:policy_params}
\end{table}

\subsection{Successes Tracking} \label{app:goal-success}
As eluded to in Section \ref{eureka-reward-design}, we use a moving-target problem formulation to facilitate continuous object rotation about a given axis. This is achieved using the keypoint position deltas outlined in Appendix \ref{app:teacher-student} to inform a success condition. Success accumulation occurs by incrementing each environment’s success tally whenever it meets the goal tolerance. Episodes are then identified for reset when they meet termination conditions or reach their length limit and the total number of resets and associated successes are recorded. The average successes per reset is computed and blended with the prior consecutive‑successes value via a fixed averaging factor. If no resets occur the prior value is maintained. Finally the updated values are averaged across all environments to yield the scalar consecutive‑successes metric, which is fed back to the LLM in reward reflection as the task score $s$, which can be seen in line 608 of Listing \ref{lst:conversation}.

\section{Compute Resources} \label{app:compute-resources}
Training for this project was conducted using a combination of a local desktop workstation and our institution's high-performance computing (HPC) cluster. The desktop machine ran Ubuntu 22.04 and was equipped with an Nvidia GeForce RTX 4090 GPU (24 GB VRAM), an Intel Core i7-13700K (13th Gen) CPU, and 64,GB RAM. On this system, up to four reward functions were trained concurrently. A full Eureka experiment, comprising five rounds of four reward function trainings, typically completed in approximately 24 hours. Full policy training to 8 billion environment steps on this setup took around 12 hours.

Additional experiments were carried out on our institution's cluster, which operates on Rocky Linux 8.9 and uses Slurm for job scheduling. Each allocated job ran on a single Nvidia Tesla P100-PCIE GPU (16 GB VRAM). Due to lower computational throughput, a full Eureka experiment on this cluster required approximately 4.5 days, and full training to 8 billion steps took around 3 days.

All experiments were conducted within a Conda-managed environment running Python 3.9, using the Isaac Gym \citep{makoviychuk2021isaac} simulation framework to leverage GPU-accelerated physics and batch policy rollouts.

\section{Real-World Objects} \label{app:real-world-objects}
Table \ref{tab:products} shows a complete list of the objects used in real-world testing.

\begin{table}[h!]
  \centering
  \scriptsize
  \begin{tabular}{@{} l c c l c c @{}}
    \toprule
    \textbf{Item} & \textbf{Dimensions (mm)} & \textbf{Mass (g)} & \textbf{Item} & \textbf{Dimensions (mm)} & \textbf{Mass (g)} \\
    \midrule
    Plastic Apple   & 75 $\times$ 75 $\times$ 70 & 60 & Tin Cylinder  & 45 $\times$ 45 $\times$ 63 & 30 \\
    Plastic Orange  & 70 $\times$ 72 $\times$ 72 & 52 & Rubber Duck   & 110 $\times$ 95 $\times$ 100 & 63 \\
    Plastic Pepper  & 61 $\times$ 68 $\times$ 65 & 10 & Gum Box       & 90 $\times$ 80 $\times$ 76 & 89 \\
    Plastic Peach   & 62 $\times$ 56 $\times$ 55 & 30 & Container     & 90 $\times$ 80 $\times$ 76 & 32 \\
    Plastic Lemon   & 52 $\times$ 52 $\times$ 65 & 33 & Rubber Toy    & 80 $\times$ 53 $\times$ 48 & 27 \\
    \bottomrule
  \end{tabular}
  \vspace{0.5em}
  \caption{Dimensions and mass of real-world objects used for testing.}
  \label{tab:products}
\end{table}

\end{document}